\documentclass[journal]{IEEEtran}

\ifCLASSINFOpdf
  \usepackage[pdftex]{graphicx}
\else
  \usepackage[dvips]{graphicx}
  \DeclareGraphicsExtensions{.eps}
\fi
\usepackage{amsmath}
\usepackage[style=ieee]{biblatex}
\usepackage[dvipsnames, table]{xcolor}
\usepackage{multirow}
\usepackage{booktabs,threeparttable}
\usepackage{diagbox}
\usepackage[graphicx]{realboxes}
\usepackage[switch]{lineno} 

\begin{document}
\title{Neuromechanical model-based adaptive control of bi-lateral ankle exoskeletons: biological joint torque and electromyogram reduction across walking conditions}
\author{Guillaume~Durandau,~\IEEEmembership{Member,~IEEE,}~Wolfgang~Rampeltshammer,~Herman~van der Kooij,~\IEEEmembership{Member,~IEEE, }and~Massimo~Sartori,~\IEEEmembership{Member,~IEEE}

\thanks{The study received funding by the European Union’s Horizon 2020 Research and Innovation Programme as part of the European Research Council (ERC) Starting Grant INTERACT (803035) and the ICT-10 Project SOPHIA (871237). The authors are with the Department of Biomechanical Engineering, Technical Medical Centre, University of Twente, Enschede 7522 NB, The Netherlands (e-mail: g.v.durandau@utwente.nl, m.sartori@utwente.nl).}}
\maketitle
\begin{abstract}


To enable the broad adoption of wearable robotic exoskeletons in medical and industrial settings, it is crucial they can adaptively support large repertoires of movements. We propose a new human-machine interface to simultaneously drive bilateral ankle exoskeletons during a range of “unseen” walking
conditions and transitions that were not used for establishing the control interface. The proposed approach used person-specific neuromechanical models to estimate biological ankle joint torques in real-time from measured electromyograms (EMGS) and joint angles. A low-level controller based on
a disturbance observer translated biological torque estimates into exoskeleton commands. We call this ”neuromechanical model-based control” (NMBC). NMBC enabled six individuals to voluntarily control a bilateral ankle exoskeleton across six walking conditions, including all intermediate transitions, i.e., two walking speeds, each performed at three ground elevations, with no need for predefined torque profiles, nor \textit{a priori} chosen neuromuscular reflex rules, or state machines as common in literature.
A single subject case-study was carried out on a dexterous locomotion tasks involving moonwalking. NMBC always enabled reducing biological ankle torques, as well as eight ankle muscle EMGs both within (22\% torque;
12\%  EMG) and between walking conditions (24\% torque;
14\% EMG) when compared to non-assisted conditions. 
Torque and EMG reductions in novel walking conditions indicated
that the exoskeleton operated symbiotically, as exomuscles
controlled by the operator’s neuromuscular system. This
opens new avenues for the systematic adoption of wearable
robots as part of out-of-the-lab medical and occupational settings.
\end{abstract}


%
\IEEEpeerreviewmaketitle
\section{Introduction}
%
%
%
%

\begin{figure*}[t]
\centering
\includegraphics[width=\textwidth]{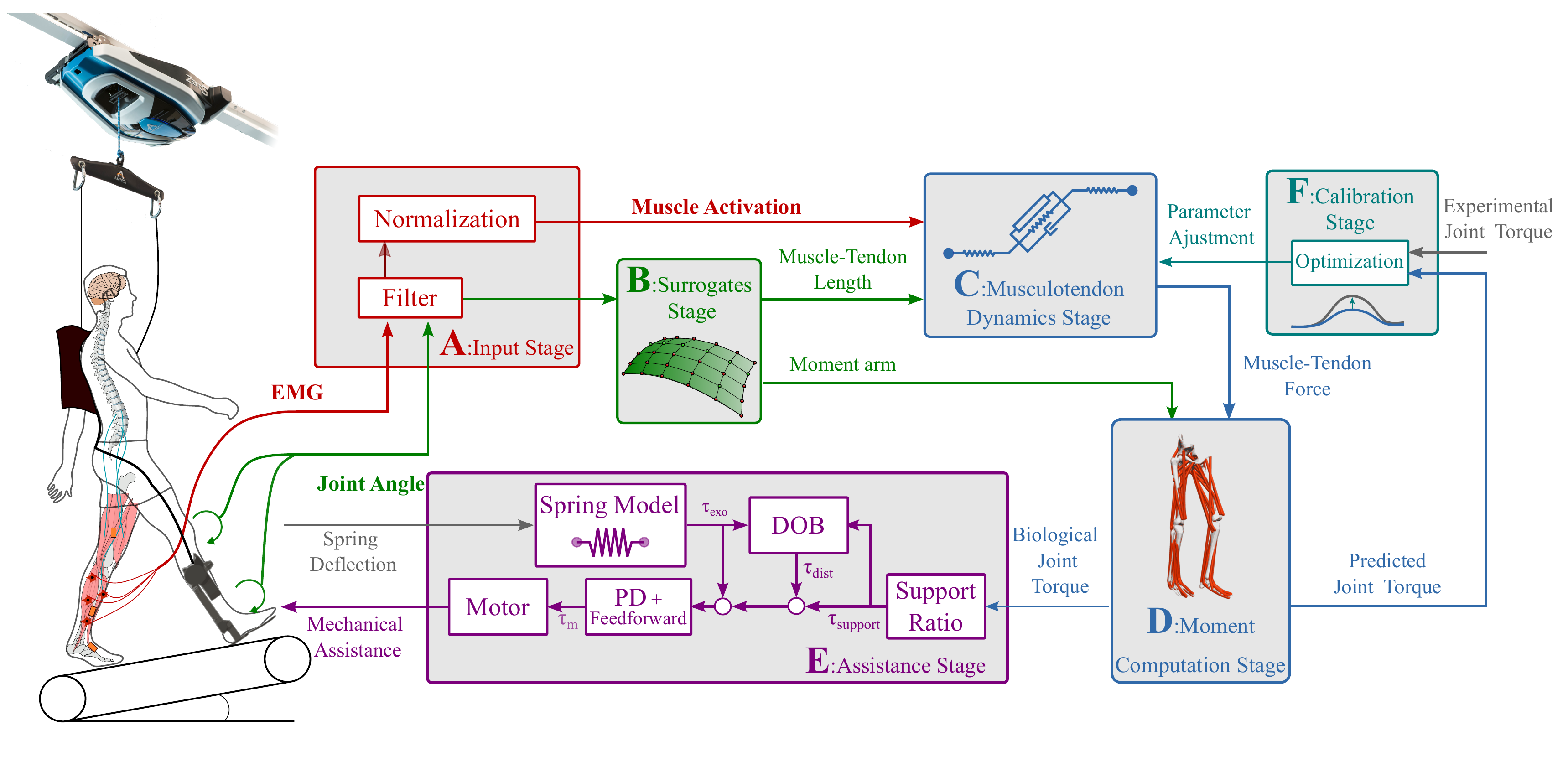}
\caption{Schematic representation of the neuromechanical
model-based controller and the experimental setup. A) input stage that records EMGs and joint angles; B) 3D musculoskeletal geometry stage that maps joint angles to muscle-tendon lengths and moment arms; C) muscle-tendon dynamics stage that transforms muscle-tendon lengths and EMGs into muscle-tendon force; D) joint moment computation stage that transforms muscle-tendon force and moment arms into net joint torque (see section II-A); E) assistance stage that transforms joint torque into mechanical assistance to the user (see section II-C, D and E) and F) the calibration stage that personalizes the muscle-tendon dynamics stage to the user (see section II-B). The experimental setup consists of the Zero-G system (on top, blue) to prevent falls, the backpack (on the back of the user, black) where the exoskeleton battery and computer unit are situated, the bilateral ankle exoskeleton (black and grey), the EMG sensors (on the leg of the user, red) and the IMU suit (on the leg of the user, orange). 
} 
\label{fig_Schema}
\end{figure*}

\IEEEPARstart{W}{EARABLE} robotic exoskeletons have great potentials for enhancing human mobility. That is, promoting motor recovery within neurorehabilitation training \cite{mehrholz2017electromechanical} or protecting from work-related musculoskeletal injuries in occupational settings (\textit{e.g.} factory or in-home scenarios) \cite{ajoudani2020smart}.
In these contexts, the ability of assisting functional movements such as walking is a central target. Walking underlies a sophisticated interplay between neurons, muscles and skeletal segments. Placing robotic exoskeletons in parallel to an already complex biological system makes the problem of effective walking assistance an open challenge \cite{Sawicki2020TheEconomy, Kim2019ReducingExosuit}. 
Advances in human-machine interfaces (HMIs) are crucial for opening up a robust and intuitive communication and control channel that enables exoskeletons to operate as a natural extension of the human neuromusculoskeletal system. Despite advances in materials \cite{Majidi2018}, soft actuation \cite{staman2018} and ergonomics \cite{Walsh2020Hinge}, wearable robotic exoskeletons are limited in their ability of seamlessly assisting a broad range of walking conditions in unforeseen and unstructured environments. Exoskeletons currently do not account for changes in walking speed, ground elevation as well as transitions across these conditions \cite{Sawicki2020TheEconomy}. Progress has been hampered by the lack of robust and intuitive HMIs. 

%
Current HMIs do not enable humans to control exoskeletons in a fully voluntary manner. Rather, they predominantly operate within \textit{a priori} defined conditions, \textit{i.e.} by relying on a finite set of pre-computed torque, angle profiles \cite{asbeck2015biologically, Galle2017ReducingPower, Kim2019ReducingExosuit, Ding2018Human-in-the-loopWalking, Khazoom2019DesignLanding, Zhang2017HumanInTheLoop} or \textit{a priori} chosen neuromuscular reflexive rules \cite{Geyer2010AActivities, Tamburella2020NeuromuscularSubjects} triggered at pre-determined gait events. Pre-defined joint torque profiles can be further optimized to individuals via human-in-the-loop optimization methods, which reduce metabolic rate \cite{Ding2018Human-in-the-loopWalking, Zhang2017HumanInTheLoop} or electromyograms (EMG) \cite{jackson2019heuristic}. However, optimized profiles are still specific to a pre-selected walking condition and require tens of minutes to be found, during which the human is exposed to sub-optimal joint torque perturbations \cite{Zhang2017HumanInTheLoop}, thereby limiting clinical viability. 
State machines are often used for switching between pre-defined profiles (\textit{i.e.} to support selected speeds and ground elevations) \cite{Kim2019ReducingExosuit,Khazoom2019DesignLanding}, but cannot provide continuous support across transitions and are prone to misclassifications and suboptimal assistance during unknown locomotion conditions.

Most experiments performed on exoskeletons do not investigate transitions across large repertoires of walking conditions \cite{Sawicki2020TheEconomy}. Although some studies investigated multiple walking speeds \cite{McCain2019MechanicsControl}, metabolic benefits were not reported during these tests. Other studies investigated different walking conditions but did not investigate the transition phase \cite{Kim2019ReducingExosuit}.

Joint angle proportional controllers, based on the difference between bi-lateral hip joint angles, showed promising results for level ground walking as well as ramp and stairs climbing \cite{Lim2019DelayedExoskeleton}. However, such methods suffer from lack of flexibility and generalizability as they are based on a two-legged inverted pendulum model, that is limited in representing walking dynamics. This limits exoskeletons applicability to pre-defined movements and prevents translation of these methods outside of the lab, ultimately impacting user's acceptance. 

Proportional myoelectric controllers \cite{McCain2019MechanicsControl,Koller2018BiomechanicsControl} have been proposed to assist the user continuously and proportionally to recorded EMGs. However, these methods rely on direct EMG control schemes where a few EMGs, typically one EMG per leg, are directly used as exoskeleton control inputs. This ignores the highly non-linear transformations that take place between EMG onset and mechanical joint torque generation due to the inherent non-linear behaviour of muscle-tendon units as well as the dynamic interplay of multiple muscles acting on one joint. Moreover, given that net joint torques are contributed by many muscles, the use of a few EMG sensors per joint, with no explicit modelling of the EMG-to-force transformations, does not enable accurate estimates of biological joint function. As a result, these methods have not shown conclusive results in terms of robustness and generalization across tasks and individuals \cite{Sawicki2020TheEconomy}.

In this study, we propose an HMI that decodes realistic estimates of ankle joint torques from leg EMGs and joint angles across a broad range of walking conditions, \textit{i.e.} two ground elevations times three walking speeds, including all transitions between these conditions. Additionally, to show the efficacy of the developed controller, a single-subject case-study was carried out involving a dexterous locomotion task, \textit{i.e.} moonwalking.
%

The proposed HMI uses a person-specific neuromechanical model of the human legs to simulate in real-time all transformations that take place from EMG onset to joint torque. This is a data-driven model-based, sensor-fusion procedure that effectively projects a higher-dimensional multi-modal set of wearable sensor signals (\textit{i.e.}, 8 leg EMGs and 4 joint angles across both legs) into a lower 2-dimensional set of ankle plantar-dorsi flexion torque profiles. We call this ”neuromechanical model-based control”, or NMBC.

This article builds on top of a previously published conference case-study \cite{durandau2020myoelectric}, which main goal was to present the use of a real-time neuromusculoskeletal model to support one walking condition on an individual healthy subject using a limited set of performance metrics. 
The present study develops the proposed NMBC to be versatile across users and walking tasks, thereby removing the need for different models for each individual walking condition. Furthermore, we present an interface between a stable and passive torque controller and an EMG-driven neuromusculoskeletal model which allows to efficiently assist walking in various conditions.
We also demonstrate beneficial biomechanical assistance on different walking conditions, including transitions between walking conditions with the same controller, with no need to re-calibrate the model, differently from state of the art techniques, which rely on other approaches (state machine or machine learning) for switching between controllers or pre-computing assistance patterns.




Methodologies based on the Hill-type muscle model were previously used to control prostheses \cite{sartori2018robust} and exoskeletons \cite{durandau2019voluntary, Cavallaro2006, Fleischer2008, 8963852, Yao2018AdaptiveModel, Ao2017MovementRobot}. Most of the previous work based on the Hill-type muscle model for lower-limb exoskeletons consisted of position tracking tasks during seated position \cite{durandau2019voluntary, Yao2018AdaptiveModel, Ao2017MovementRobot} or slow locomotion tasks with single joint support \cite{Fleischer2008}. None of the previous studies has applied EMG-driven neuromusculoskeletal modelling to control bilateral exoskeletons and support walking across a large repertoire of conditions and transitions. 


%
%
NMBC derives muscle activations and resulting mechanical forces from measured EMGs. This allows making no assumption on how muscle activate, \textit{i.e.} there is no need to choose finite sets of muscle reflex rules or optimal activation criteria that are task dependent \cite{Kim2019ReducingExosuit}. 

We hypothesized that NMBC calibration procedure and integration with an exoskeleton low-level torque controller with apparent passivity would enable estimating realistic joint torque across a broad range of "unseen" walking conditions, \textit{i.e.} walking conditions that were not used to calibrate the model. Moreover, we hypothesised that, when used to control bi-lateral exoskeleton, NMBC would lead to biological joint torque and EMG reduction across different unseen conditions. From this, we can formulate two research questions: (I) can biological joint torque amplitude be reduced when assisted by NMBC-controlled exoskeleton for all tested locomotion conditions and transitions (primary outcome)?; (II) Can EMG amplitude be reduced when assisted by NMBC-controlled exoskeleton for all tested locomotion conditions and transitions (secondary outcome)?


We first describe the structure of NMBC as well as that of the wearable bilateral ankle exoskeleton employed in this study. Then, we present the experimental procedures for testing the efficacy of our proposed approach as well as the quantitative analyses and results emerging from the performed experiments. Finally, we discuss the implications and limitations of the study and future work.


\section{Methods}
Fig. 1 depicts our proposed NMBC scheme. Each component is detailed in the remainder of this section.

\begin{figure}[t]
\centering
\includegraphics[width=\linewidth]{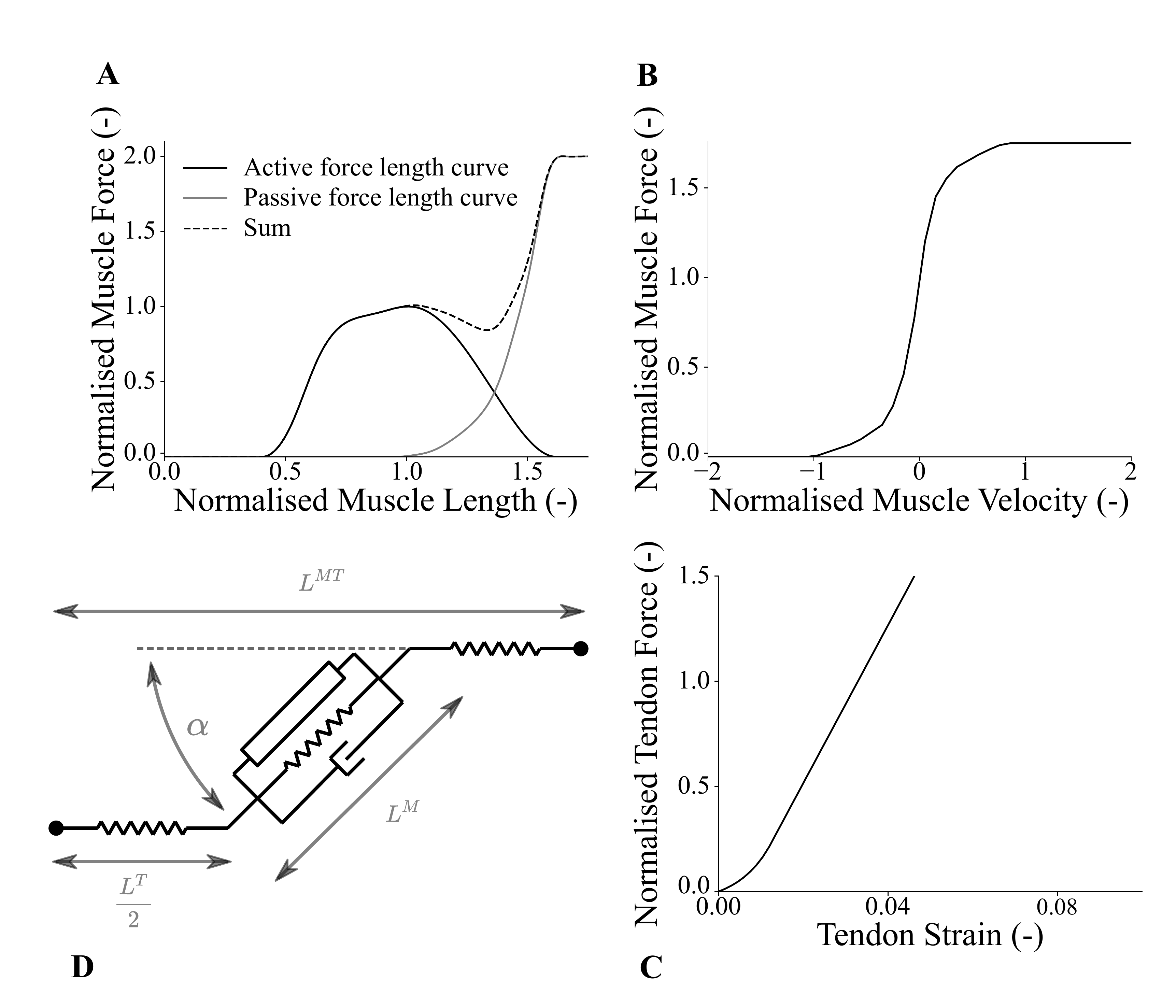}
\caption{Hill-type muscle model schematics along with the normalized force-length-velocity relationships used within NMBC. A) normalized muscle fiber's active force-length and passive force-length relationships, B) normalized muscle fiber force-velocity relationship, C) elastic tendon force-strain relationship D) Hill-type muscle representation.}
\label{fig_Hill}
\end{figure}

\subsection{High-level control via NMBC}
\label{subsec_NMSModel}
To assure voluntary and continuous torque control, NMBC computes exoskeleton commands as a direct function of the subject's estimated biological joint torque. EMGs are recorded, amplified and filtered via hardware directly by the surface electrodes using proprietary signal detection and acquisition system (AxonMaster 13E500, Ottobock, Germany) (Fig. \ref{fig_Schema}-A, input Stage). Filtered EMGs are normalized using pre-recorded maximal voluntary contractions to compute muscle excitation. The tasks used for maximal voluntary contraction recording are static co-contraction of the plantar and dorsiflexor muscles maintained for a few seconds as well as dynamic calf and front foot rises. Each task was repeated three times. 
Filtered and normalized EMGs are subsequently converted into muscle activations (see appendix \ref{ann::muscleActivation}) and used to drive a set of virtual muscle-tendon units (see Section III for the definitions of the muscle-tendon units and EMGs used). 

To update the kinematic-dependent states of the virtual muscle-tendon units, the muscle-tendon operating length needs to be known. Muscle-tendon kinematics cannot be easily recorded directly but can be computed as a function of joint angles using three-dimensional musculoskeletal geometry models. To assure a real-time estimation of muscle-tendon length, we employed B-splines as previously presented \cite{sartori2012estimation}. This allows for the computation of muscle-tendon lengths and moment arms as a function of joint angles. B-spline coefficients for each muscle-tendon unit are computed using nominal length values generated via OpenSim \cite{delp2007opensim}.
Muscle-tendon units' length and activation estimates are used to compute the resulting muscle-tendon force via personalized Hill-type muscle models (Fig. \ref{fig_Schema}-C, Musculotendon dynamics stage) (see appendix \ref{ann::MTU}). These consist of a non-linear spring (tendon) in series with three elements representing the muscle fibers: an active contractile element, in parallel with a non-linear spring and a linear damper (\textit{i.e.} muscle fiber passive elements) \cite{sartori2011emg}.


From joint angles the whole length of an individual muscle-tendon unit is obtained but the constituent muscle fiber length is needed to compute muscle fiber force. For this, the Brent–Dekker root-solver iterative method \cite{brent1973some} is used to solve for the equilibrium between muscle force and tendon force. Tendon force is obtained using the passive tendon force-strain relationship (fig. \ref{fig_Hill}-C) (see appendix \ref{ann::MT}). 

Finally, muscle forces are projected onto the ankle joint plantar-dorsiflexion via the moment arms to obtain joint torque. The moment arm is obtained via the partial derivative relative to a joint angle using the B-spline algorithm previously introduced and the principle of virtual work \cite{sartori2012estimation} (see appendix \ref{ann::MA}).


\subsection{Model personalization} \label{subsec_personalization}
A generic musculoskeletal geometry model is scaled linearly to each individual using the open-source software OpenSim \cite{delp2007opensim} and 3D motion capture data of body landmarks (bony areas) recorded during a static standing pose. During this procedure, we linearly adjust muscle-tendon bone-wrapping and origin insertion points as well as the center of mass values and positions of the anatomical segments, to match an individual's anthropometry. This scaled model is used to create a multidimensional B-Spline function per muscle-tendon unit as described in Section \ref{subsec_NMSModel} for the computation of subject-specific muscle tendon lengths and moment arms.
Four parameters are calibrated for each muscle-tendon unit in the model including: $E$ the EMG shape factor, $L^{T}_{Slack}$ the tendon slack length, $L^{M}_{Opt}$ the optimal fiber length and $F^{Max}_{Iso}$ the muscle maximal isometric force. This calibration is based on a two steps procedure. First, a previously presented pre-tuning procedure \cite{winby2008evaluation} is employed to identify initial values for $L^{M}_{Opt}$ and $L^{T}_{Slack}$ (see appendix \ref{ann::pre-calib}). 

After initial values for $L^{M}_{Opt}$ and $L^{T}_{Slack}$ are found, all four muscle parameters ($E$, $F^{Max}_{Iso}$, $L^{M}_{Opt}$, and $L^{T}_{Slack}$) are further optimized to enable the subject-specific model to fuse recorded EMGs and joint angles into joint torque profiles over a range of locomotion trials. This is based on a simulated annealing procedure \cite{goffe1994global} that minimizes the squared error between the estimated torque by the model and the experimental torque derived via inverse dynamics (Fig. \ref{fig_Schema}-F) (see appendix \ref{ann::calib}).
The parameters boundaries varied between -3 and 0 for the EMG shape factors $E$, between 0.5 and 1.5 for the normalized maximal isometric muscle forces $F^{Max}_{Iso}$ after normalization and between $\pm2.5\%$ for the optimal fiber lengths $L^{M}_{Opt}$ and between $\pm5\%$ tendon slack lengths $L^{T}_{Slack}$ from the values found during pre-tuning.

\subsection{Bi-lateral ankle exoskeleton}
In this study, the left and right ankle modules of the Symbitron exoskeleton with upgraded electronics were used~\cite{Meijneke2021SymbitronIndividuals} to assist bi-lateral plantar-dorsiflexion during locomotion. Each ankle module weighs 5Kg. The active degree of freedom is actuated with a rotary series elastic actuator~(SEA), which transmits the desired interaction forces via a push-pull rod from its distal location to the ankle joint. The SEA consists of a motor (Tiger Motor U8-10(Pro), T-Motor, Nancheng, China) that is connected to a harmonic drive (LCSG20, Leader Drive, Jiangsu, China) with a gear ratio of 1:100. The harmonic drive is connected to the output of the motor with a custom rotary spring with a stiffness of 1534 Nm/rad. The actuator can deliver a controlled peak torque of 100 Nm and has a maximum output speed of 5 rad/s. The motor is controlled via an Everest Net drive (Ingenia, Barcelona, Spain), which communicates with the control computer via EtherCAT. Motor position is measured via a rotational encoder (16 b MHM, IC Haus, Bodenheim, Germany). Additionally, the actuator measures the spring deflection and joint position with two encoders (20 b Aksim, RLS (Renishaw), Kemnda, Slovenia) which are transmitted to the control computer via the Everest Net drive. The backpack contains the control computer, a NUC (Intel, Santa Clara, USA) that executes the controller in TwinCAT 3 (Beckhoff Automation, Verl, Germany) in real-time with a sampling frequency of 1 kHz. Additionally, the backpack contains two batteries, supplying the computer and actuators with power. The backpack has a weight of 10 kg.

\subsection{Low-level torque control via disturbance observers}
The Symbitron ankle exoskeleton interacts with its user by controlling the interaction torque between user's leg and the exoskeleton module, \textit{i.e.} the spring torque $\tau_{exo}$. This torque is computed from the measured spring deflection and a linear model of the actuator's spring (Fig. \ref{fig_Schema}-E). The interaction torque is controlled via a disturbance observer controller recently developed for lower limb exoskeletons \cite{Rampeltshammer2020AnActuators}. 
The controller fulfils three purposes: it increases the bandwidth of the actuator to 30~Hz, it lowers the actuator's apparent impedance, \textit{i.e.} it makes the actuator as mechanically transparent as possible, and it guarantees unconditional interaction stability with any environment. Unconditional interaction stability is especially important for an ankle exoskeleton to avoid non-passive behavior, \textit{i.e.} oscillations, during impacts such as heel strikes and changes in the environment such as between swing and stance phase. This unconditional interaction stability was achieved by adapting the disturbance observer, and has been proven with the actuators used in the Symbitron exoskeleton in \cite{Meijneke2021SymbitronIndividuals}. It was found that the low-level torque controller was able to achieve a precise torque tracking with a root mean squared error of 1.8 N.m.

The controller consists of an inner loop PD controller with feedforward term, that increases the actuator's torque bandwidth, and an outer loop disturbance observer (DOB) that lowers the actuator's apparent impedance, while keeping it passive. The DOB computes the torque caused by disturbances, such as impacts during heel strike, or voluntary motion of the subject, and subtracts that disturbance torque $\tau_{dist}$ from the desired reference torque $\tau_{support}$, which is the estimated joint torque from the NMBC multiplied by a support ratio, to eliminate the effect of the disturbance on the interaction torque $\tau_{exo}$. This disturbance rejection makes the actuator as transparent as possible while keeping its interaction with the environment stable. The resulting torque $\tau_m$ is sent to the motor as a reference.

\subsection{Assistance}
To assure timely and voluntary exoskeleton torque control a tight integration between high and low-level control is required. For this, the estimated biological joint torque is sent from the subject-specific NMBC to the exoskeleton low-level controller via the Ethercat real-time communication protocol. The low-level controller multiplies biological torque estimates by a support ratio that varies between 0 and 1, with 0 meaning that the exoskeleton acts in minimal impedance and 1 signifying 100\% of assistance given, \textit{i.e.} the exoskeleton assists with the same amount of torque as generated by the subject's biological joint. The maximal torque delivered by the exoskeleton is limited to a maximum of 40 N.m to assure the safety of the subject and the integrity of the actuator.

\subsection{Data processing}
Experimental results were segmented automatically using a peak detector on the knee joint angle. Each segment was re-sampled as percentage of the gait cycle. The root means squared value of the data of interest (EMG, biological torque, exoskeleton torque, joint angle) was identified for each segmented cycle, right and left side results were averaged, values that were larger than three times the interquartile values were removed. Moreover, percentages of change between assisted and non assisted condition were calculated using the mean of all steps for EMG and joint torques (biological and exoskeleton). All results presented in this article are for the average of left and right leg.

Statistical significance was computed for the percentage of change between exoskeleton conditions. First, normality of the distribution was verified using the Shapiro–Wilk test. Then, the equality in variance was verified using the Levene's test. If both tests were passed a two-sided t-test was used to compute significance. If inequality of variance was found the Welch’s t-test was used. This was the case for most of the locomotion conditions but for some of the locomotion conditions, neither of the assumptions were true and the Wilcoxon signed-rank test was used. 
For the locomotion condition transitions, significance was computed using linear mixed modelling with the torque or individual muscle's EMG as a fixed effect, the transition as an independent variable, and exoskeleton conditions as a random effect. For this test, a Bonferroni correction was used to reduce the effect of multi-testing (locomotion conditions as well as transitions) for the torque reductions and each muscle's EMG (P \textless 0.007).

\begin{figure}[t]
\centering
\includegraphics[width=0.5\textwidth]{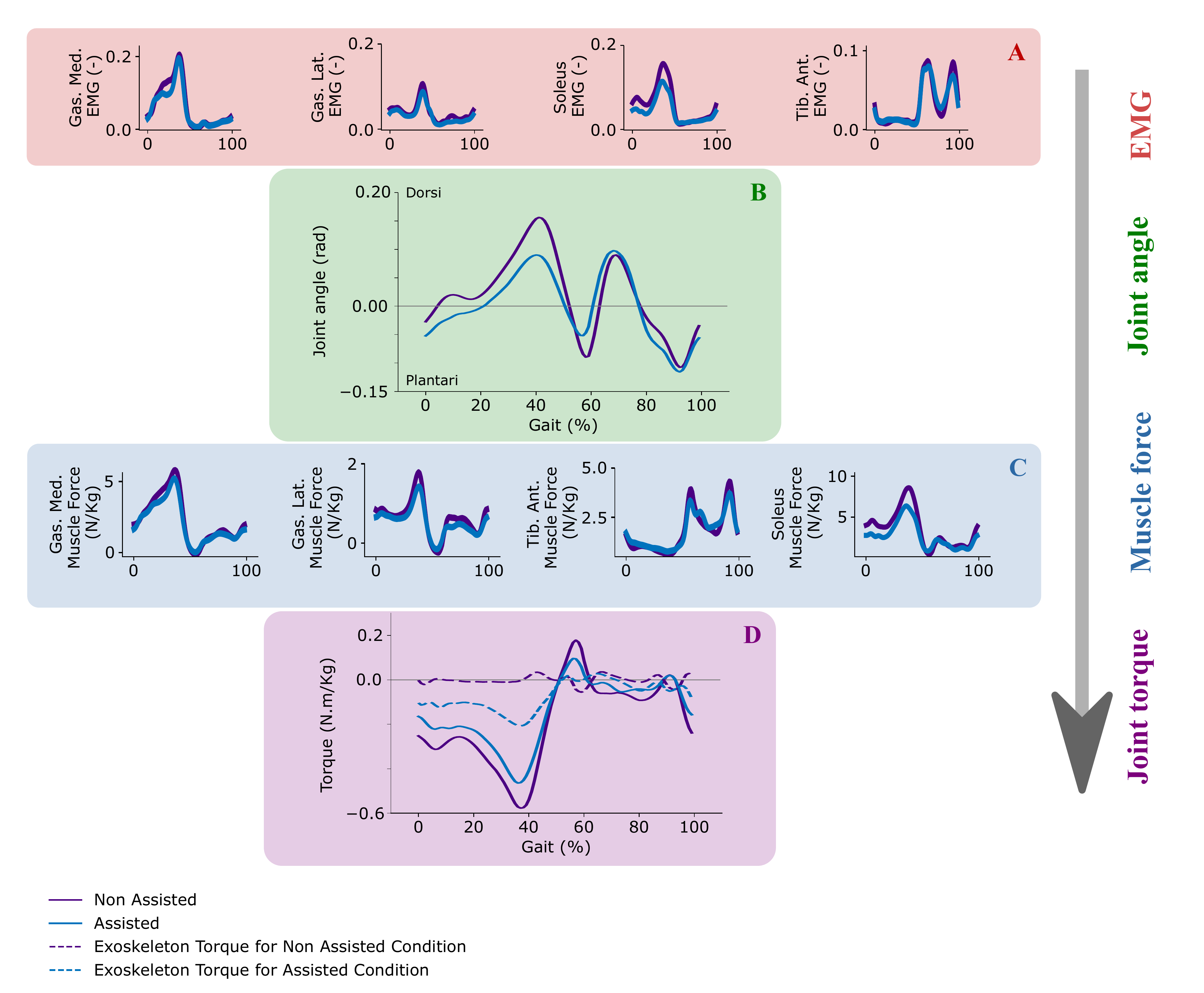}
\caption{Averaged EMGs for the Gastrocnemius Medialis, Gastrocnemius Lateralis, Soleus and Tibialis Anterior (A - red background), ankle joint angle (B - green background), muscle force for the Gastrocnemius Medialis, Gastrocnemius Lateralis, Soleus and Tibialis Anterior (C - blue background), plantar dorsiflexion biological joint torque and exoskeleton torque (D - purple background) over the gait cycle for the 0.5 m/s, -5\% elevation walking condition. The represented data are averaged across all subjects. }
\label{fig_all}
\end{figure}

\begin{figure*}[t]
\centering
\includegraphics[width=\textwidth]{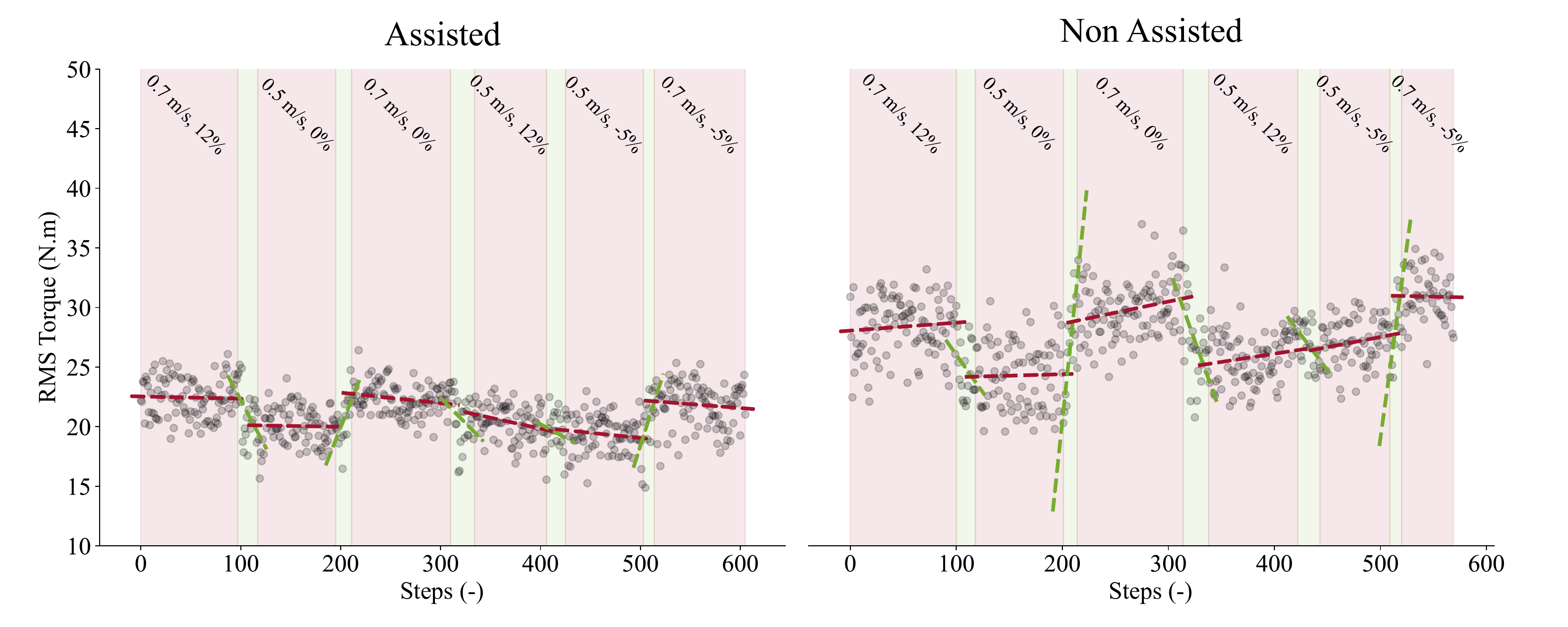}
\caption{Estimated biological ankle torque for a representative subject for the two tested locomotion conditions, \textit{i.e.} assisted and non-assisted. Each grey dot represents the joint torque root mean squared sum (RMS) for each gait cycle. The red dotted line represents joint torque RMS trend within each locomotion condition. The green dotted line represents torque RMS trend during transitions across locomotion conditions.}
\label{fig_adpatation}
\end{figure*}

\begin{figure*}[t]
\centering
\includegraphics[width=\textwidth]{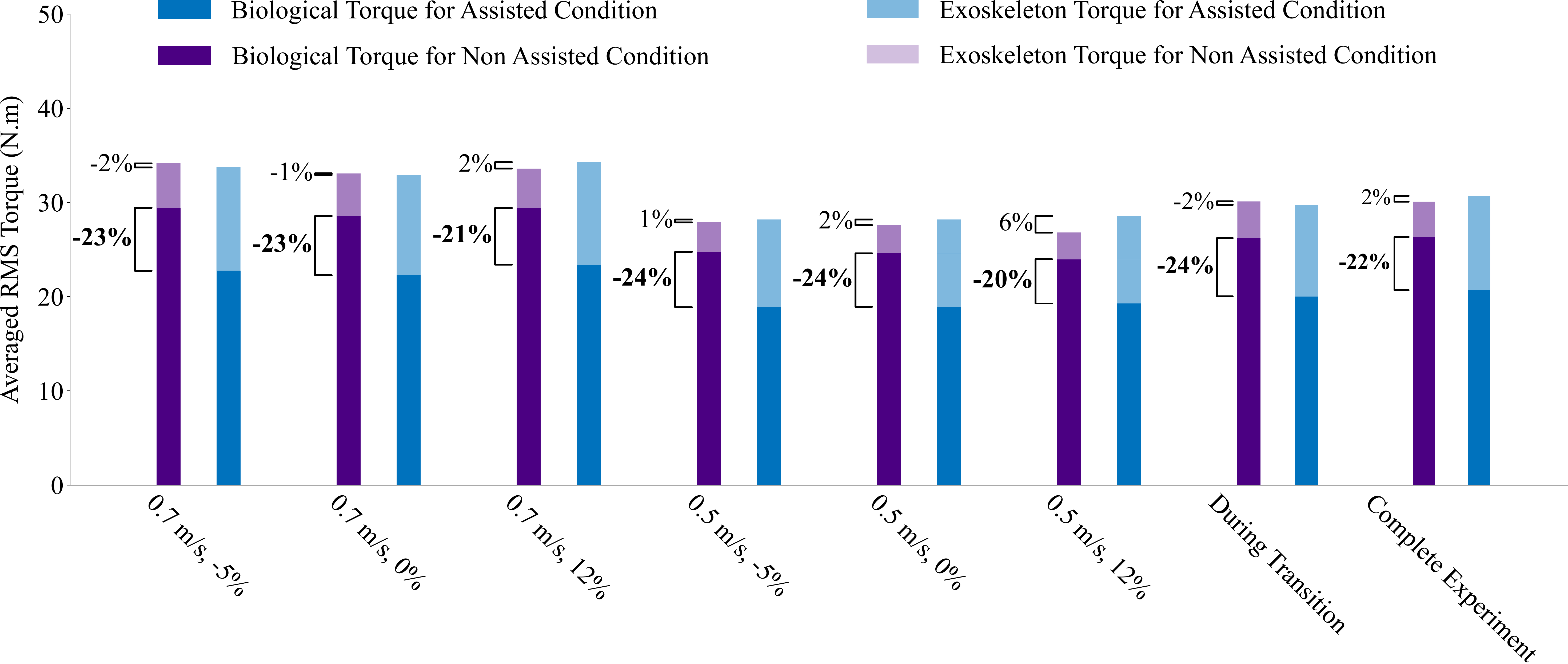}
\caption{Ankle joint torque root mean squared sum averaged across all subjects within each tested locomotion condition, \textit{i.e.} assisted (in blue) and non-assisted (minimal impedance, in purple). The vertical bar darker shade represents the human biological torque estimated via NMBC. The vertical bar lighter shade represents the exoskeleton assistive torque as recorded via the actuator spring deflection (which are added on top of the biological joint torque). The number on top of each vertical bar represents the difference in total ankle torque (exoskeleton plus biological torque) between assisted and non-assisted exoskeleton conditions. The top lower number in bold represents the reduction in biological torque between assisted and non-assisted exoskeleton conditions. During transition represents the average of all transition happening between two locomotion conditions (\textit{i.e.} change in speed, elevation or both) and the Complete experiment represents the full recorded experiment with locomotion conditions and locomotion conditions transition.}
\label{fig_torque_bar}
\end{figure*}

\begin{figure}[t]
\centering
\includegraphics[width=0.5\textwidth]
{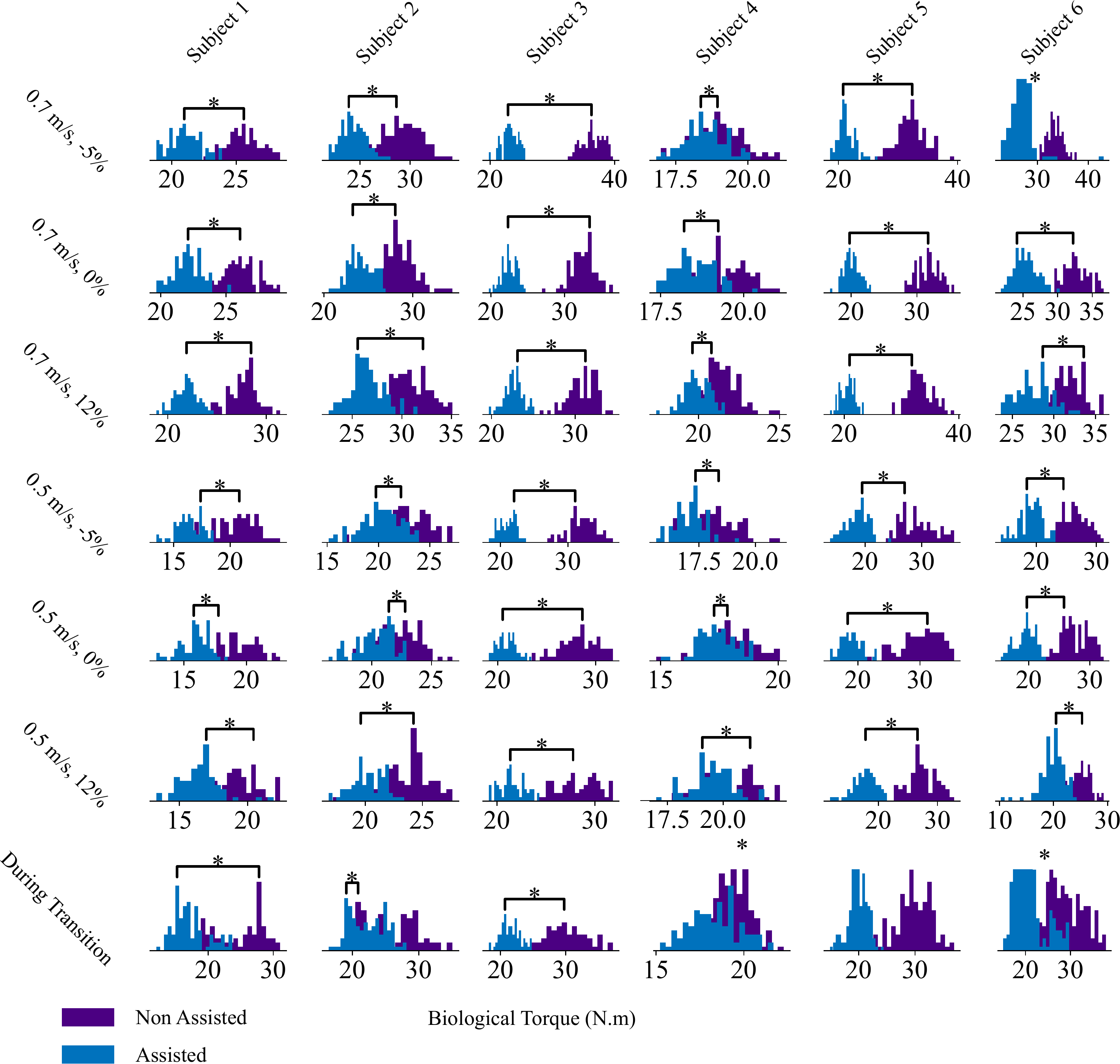}
\caption{Distribution of biological torque amplitude, \textit{i.e,} number of occurrences of the averaged ankle biological joint torque for each gait cycle for the two tested conditions (non assisted walking (purple) and assisted walking (blue)) and all tested locomotion conditions as well as locomotion conditions transition. The symbol '*' represents statistical significance (P \textless 0.007).}
\label{fig_torque_desnity}
\end{figure}

\begin{figure}[t]
\centering
\includegraphics[width=0.5\textwidth]{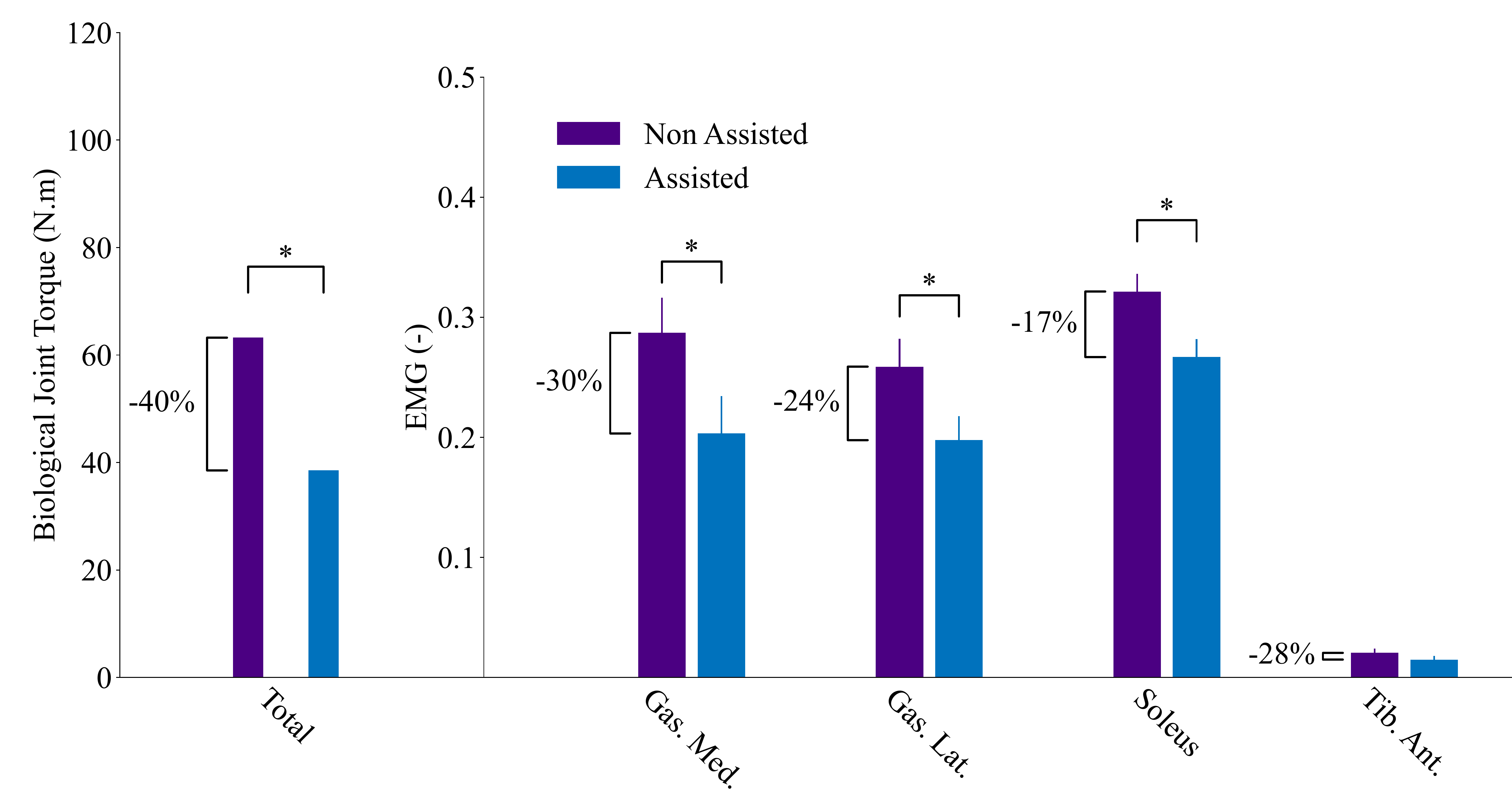}
\caption{EMG reduction for all muscles and ankle biological joint torque reduction during the moonwalk task. Data are reported for the assisted (blue) and non-assisted (purple) exoskeleton conditions. The symbol ’*’ represents statistical significance (P \textless 0.05).} 
\label{fig_EMG_moonwalk}
\end{figure}

\begin{figure*}[t]
\centering
\includegraphics[width=\textwidth]{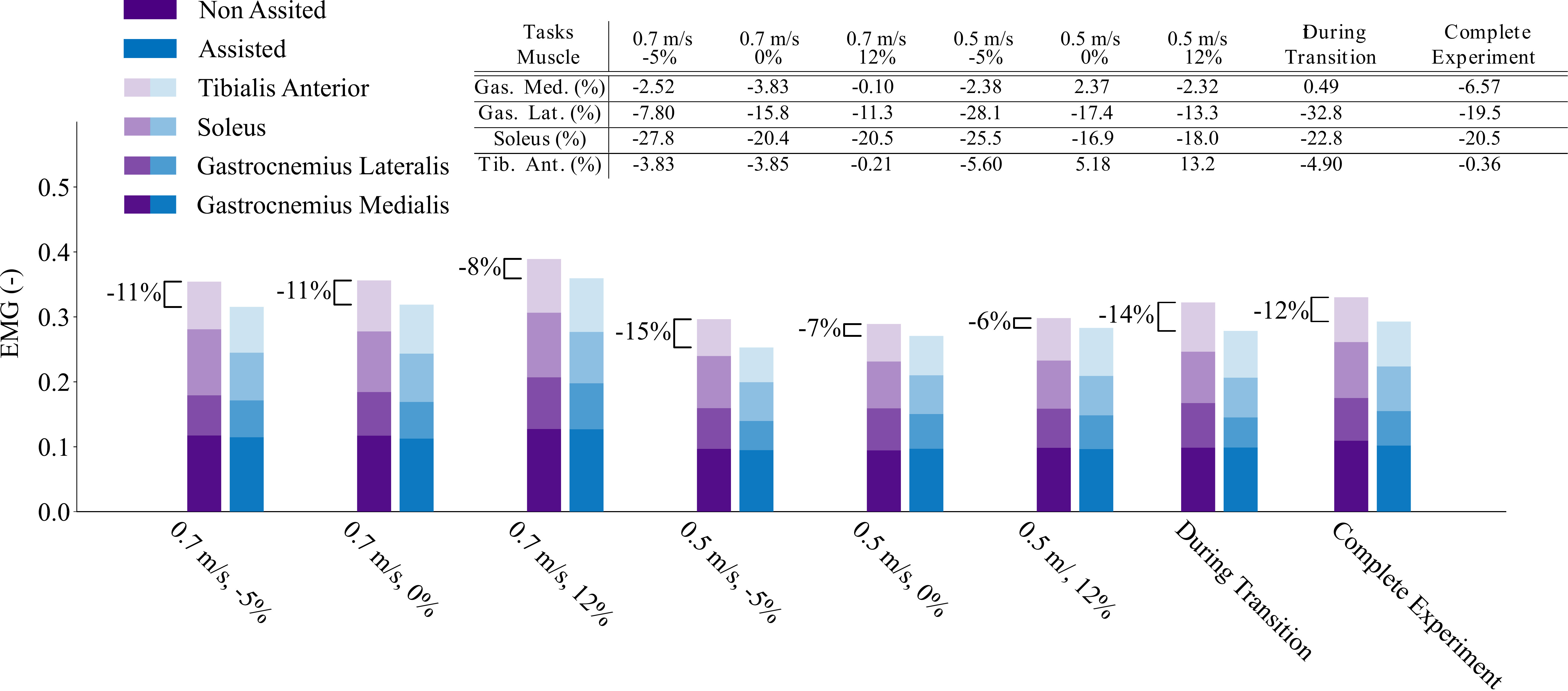}
\caption{EMG reduction across all subjects within each walking condition as well as during transitions and the entire experiment (\textit{i.e.} all locomotion conditions and transitions altogether). Data are reported for the assisted (blue) and non-assisted (purple) exoskeleton condition. During transition represents the average of all transition happening between two locomotion conditions (\textit{i.e.} change in speed, elevation or both). Muscle-specific EMG reduction results across all locomotion conditions, transitions and complete experiments are presented in the enclosed table. }
\label{fig_EMG}
\end{figure*}

\section{Experiments}
Experiments were conducted on 6 healthy subjects (28$\pm$5 years, 176$\pm$7 cm, 73$\pm$12 Kg, 5 male/1 female). Participants had no instance of musculoskeletal injury or motor-control impairment, had a shoe size between 43 (9.5 US) and 46 (12 US) and a tibial vertical length greater than 28 cm.
All experimental procedures were carried out in accordance with the Declaration of Helsinki on research involving human subjects. All subjects provided their explicit written informed consent to participate. Experiments were approved by the Natural Sciences and Engineering Sciences Ethics committee of the University of Twente (reference number 2020.21).

The experiment was conducted following three phases each one done on a different day. The experiment first phase was carried out for the purpose of personalizing NMBC to each individual. Motion capture's 3D markers data (Oqus, Qualisys, Sweden), three-dimensional ground reaction forces (M-Gait, MotekForce Link, The Netherlands) and EMGs (AxonMaster 13E500, Ottobock, Germany) were recorded from each subject. EMG signals from eight muscles were recorded including: left and right Soleus, Tibialis Anterior, Gastrocnemius Medialis and Lateralis. Marker data and ground reaction forces were used to compute joint angles and joint torques using inverse kinematics and inverse dynamics \cite{delp2007opensim}. These data were used to personalize NMBC using the methods described in section \ref{subsec_personalization}. NMBC's musculoskeletal model used during this experiment was based on \cite{delp1990interactive} and contained the following joints: left and right plantar dorsiflexion and knee flexion-extension and the following 14 muscles tendon units: left and right Soleus, Tibialis Anterior, Gastrocnemius  Medialis and Lateralis and Peroneus Longus, Brevis and Tertius. The following tasks were used for the calibration of the model: static pose for 10 sec, 60 seconds of treadmill walking at 0.5 m/s and 0.7 m/s, 10 calf rises and 10 front foot rises.

The second phase of the experiment consisted of defining the locomotion experimental parameters and acclimation of the subject to walking with the exoskeleton. Two experimental parameters were set during this phase, the support ratio (see Table \ref{table:1}) and the step frequency for the two tested speeds. Step frequency (the time between two hell strike from the same leg) was controlled during the experiment to make sure that the assistance was integrated by the subjects and reduced their own torque without altering their speed or step length. 

The support ratio (see section II-E) was selected by each subject as a tradeoff between comfort and assistance. Since most of our subjects were exoskeleton naïve users, comfort level across users was different. We privileged subjects' comfort over a larger support ratio. 

Subjects walked on a treadmill (Thera-Treadpro, Sportplus) with the exoskeleton with and without assistance until they presented a natural-looking gait and felt confident (10 to 20 minutes on average). To measure the knee joint angle, an IMU suit was used (Link, Xsens, the Netherlands), the ankle joint angle was directly available from the joint encoder of the exoskeleton. The knee angles are required for the simulated Gastrocnemius muscles that span the ankle and knee joints. A fall prevention system was used every time the subjects were walking with the exoskeleton (ZeroG, Aretech LLC, USA), which provided a body-weight support of 5 Kg.

The last phase of the experiment evaluated the NMBC's ability in supporting locomotion and exoskeleton voluntary control (see Video in supplementary material). Two exoskeleton conditions were tested, non assisted (minimal impedance) and assisted (NMBC-based). Each exoskeleton condition was tested across six different locomotion conditions that were randomly presented to the subject. The data were collected in a single uninterrupted session including the six randomly combined locomotion conditions and the transitions within. The tested locomotion conditions were 0.5 m/s (speed), 0\% (0 degrees) (slope); 0.5 m/s, -5\% (-2.8 degrees); 0.5 m/s, 12\% (6.84 degrees); 0.7 m/s 0\% (0 degrees), 0.7 m/s, -5\% (-2.8 degrees); 0.7 m/s 12\% (6.84 degrees).

Each locomotion condition had a duration of 3 min, the transition between walking conditions was of variable length due to the time needed by the treadmill to change between speed and/or inclination. These varied approximately between 10 seconds and 60 seconds. 

For one of the subject (subject 5), a case-study was conducted, which involved moonwalking. The task was realized at the end of the experiment and was repeated 3 times for each condition on a 5 m track (see video in supplementary material.)

\begin{table*}
\begin{center}
 \caption{Subject specific support ratio, EMG and torque reduction as well as averaged received peak assistance torque for all tasks, transition and complete experiments. Red cell represents increase and blue cell represents reduction.}
\label{table:1}
 \renewcommand*\TPTnoteLabel[1]{\parbox[b]{3em}{\hfill#1\,}}
 \begin{threeparttable}
\begin{tabular}{c | c | c c c c c c c c|} 
Subject (-)  & \multirow{2}{*}{\backslashbox{Muscles}{Tasks}} &0.7 m/s& 0.7 m/s & 0.7 m/s& 0.5 m/s&0.5 m/s&0.5 m/s&During & Complete \\
Support Ratio (\%)& &-5\%&0\%&12\%&-5\%&0\%&12\%&Transition&Experiment\\
\hline
\hline
\multirow{12}{*}{Subject 1 (60\%)}&\multirow{2}{*}{Gas. Med. (\%)} &\cellcolor{blue!20} &\cellcolor{blue!20} &\cellcolor{blue!20} &\cellcolor{blue!20} & \cellcolor{blue!20}&\cellcolor{blue!20} &\cellcolor{blue!20} &\cellcolor{blue!20} \\
 & &\cellcolor{blue!20} \multirow{-2}{*}{-14.6*}&\cellcolor{blue!20}\multirow{-2}{*}{-9.8*} & \cellcolor{blue!20}\multirow{-2}{*}{-8.6*}& \cellcolor{blue!20}\multirow{-2}{*}{-17.1*}&\cellcolor{blue!20}\multirow{-2}{*}{-3.6} &\cellcolor{blue!20}\multirow{-2}{*}{-11.1*} &\cellcolor{blue!20}\multirow{-2}{*}{-10.2*} &\cellcolor{blue!20} \multirow{-2}{*}{-10.0}\\
\cline{2-10}
 &\multirow{2}{*}{Gas. Lat. (\%)} &\cellcolor{blue!20} & \cellcolor{blue!20}& \cellcolor{red!20}& \cellcolor{blue!20}& \cellcolor{blue!20}&\cellcolor{blue!20} &\cellcolor{blue!20} &\cellcolor{blue!20} \\
& &\cellcolor{blue!20} \multirow{-2}{*}{-20.6*}&\cellcolor{blue!20}\multirow{-2}{*}{-15.2*} &\cellcolor{red!20}\multirow{-2}{*}{9.2*} &\cellcolor{blue!20}\multirow{-2}{*}{-10.3} &\cellcolor{blue!20}\multirow{-2}{*}{-0.8} &\cellcolor{blue!20}\multirow{-2}{*}{-6.7} &\cellcolor{blue!20}\multirow{-2}{*}{-2.9} &\cellcolor{blue!20} \multirow{-2}{*}{-8.5}\\
\cline{2-10}
&\multirow{2}{*}{Soleus (\%)} &\cellcolor{blue!20} &\cellcolor{blue!20} &\cellcolor{blue!20} &\cellcolor{blue!20} &\cellcolor{blue!20} & \cellcolor{blue!20}& \cellcolor{blue!20}&\cellcolor{blue!20} \\
& &\cellcolor{blue!20} \multirow{-2}{*}{-40.6*}&\cellcolor{blue!20}\multirow{-2}{*}{-27.4*} &\cellcolor{blue!20} \cellcolor{blue!20}\multirow{-2}{*}{-22.6*}& \cellcolor{blue!20}\multirow{-2}{*}{-37.0*}&\cellcolor{blue!20}\multirow{-2}{*}{-17.2*} & \cellcolor{blue!20}\multirow{-2}{*}{-9.5*}&\cellcolor{blue!20} \multirow{-2}{*}{-22.1*}& \cellcolor{blue!20}\multirow{-2}{*}{-27.7}\\
\cline{2-10}
&\multirow{2}{*}{Tib. Ant. (\%)} &\cellcolor{red!20} & \cellcolor{red!20}&\cellcolor{red!20} &\cellcolor{red!20} & \cellcolor{red!20}& \cellcolor{red!20}& \cellcolor{red!20}& \cellcolor{red!20}\\
& &\cellcolor{red!20}\multirow{-2}{*}{13.5*} &\cellcolor{red!20}\multirow{-2}{*}{12.2*} &\cellcolor{red!20}\multirow{-2}{*}{35.8*} & \cellcolor{red!20}\multirow{-2}{*}{9.3}&\cellcolor{red!20}\multirow{-2}{*}{43.6*} & \cellcolor{red!20}\multirow{-2}{*}{32.4*}&\cellcolor{red!20}\multirow{-2}{*}{16.8} &\cellcolor{red!20}\multirow{-2}{*}{18.8} \\
\cline{2-10}
&Biological &\cellcolor{blue!20} & \cellcolor{blue!20}&\cellcolor{blue!20} &\cellcolor{blue!20} &\cellcolor{blue!20} &\cellcolor{blue!20} &\cellcolor{blue!20} & \cellcolor{blue!20}\\
&Torque (\%) &\cellcolor{blue!20}\multirow{-2}{*}{-36.2*} & \cellcolor{blue!20}\multirow{-2}{*}{-30.8*}&\cellcolor{blue!20} \multirow{-2}{*}{-26.7}&\cellcolor{blue!20}\multirow{-2}{*}{-34.3*} &\cellcolor{blue!20}\multirow{-2}{*}{-25.9*} &\cellcolor{blue!20}\multirow{-2}{*}{-23.6*} &\cellcolor{blue!20}\multirow{-2}{*}{-28.0*} & \cellcolor{blue!20} \multirow{-2}{*}{-28.6}\\
\cline{2-10}

&Averaged & & & & & & & & \\
& Peak Assistance (N.m)&\multirow{-2}{*}{33.8}&\multirow{-2}{*}{36.5} &\multirow{-2}{*}{37.5} &\multirow{-2}{*}{12.8} &\multirow{-2}{*}{32.3} &\multirow{-2}{*}{34.0} &\multirow{-2}{*}{34.3} & \multirow{-2}{*}{35.1}\\
\hline
\multirow{12}{*}{Subject 2 (40\%)}&\multirow{2}{*}{Gas. Med. (\%)} &\cellcolor{red!20} &\cellcolor{blue!20} &\cellcolor{red!20}& \cellcolor{red!20}& \cellcolor{red!20}&\cellcolor{blue!20} &\cellcolor{red!20} & \cellcolor{blue!20} \\
 & &\cellcolor{red!20} \multirow{-2}{*}{0.2}&\cellcolor{blue!20}\multirow{-2}{*}{-4.2} &\cellcolor{red!20}\multirow{-2}{*}{7.5*} &\cellcolor{red!20}\multirow{-2}{*}{5.5} &\cellcolor{red!20}\multirow{-2}{*}{3.4} & \cellcolor{blue!20}\multirow{-2}{*}{-5.2}&\cellcolor{red!20}\multirow{-2}{*}{5.1} &\cellcolor{blue!20} \multirow{-2}{*}{-0.5}\\
\cline{2-10}
 &\multirow{2}{*}{Gas. Lat. (\%)} &\cellcolor{red!20} &\cellcolor{blue!20} &\cellcolor{red!20} & \cellcolor{red!20}& \cellcolor{red!20}& \cellcolor{red!20}&\cellcolor{blue!20} &\cellcolor{red!20} \\
& & \cellcolor{red!20}\multirow{-2}{*}{4.5}&\cellcolor{blue!20}\multirow{-2}{*}{-6.8} &\cellcolor{red!20}\multirow{-2}{*}{26.4*} &\cellcolor{red!20}\multirow{-2}{*}{2.0} & \cellcolor{red!20}\multirow{-2}{*}{2.3*}& \cellcolor{red!20}\multirow{-2}{*}{1.5}&\cellcolor{blue!20}\multirow{-2}{*}{-16.6} & \cellcolor{red!20}\multirow{-2}{*}{2.0}\\
\cline{2-10}
&\multirow{2}{*}{Soleus (\%)} &\cellcolor{blue!20} &\cellcolor{blue!20} &\cellcolor{blue!20} &\cellcolor{blue!20} &\cellcolor{blue!20} &\cellcolor{blue!20} & \cellcolor{blue!20}& \cellcolor{blue!20}\\
& & \cellcolor{blue!20}\multirow{-2}{*}{-35.0*}& \cellcolor{blue!20}\multirow{-2}{*}{-23.6*}& \cellcolor{blue!20}\multirow{-2}{*}{-27.0*}&\cellcolor{blue!20}\multirow{-2}{*}{-27.6*} & \cellcolor{blue!20}\multirow{-2}{*}{-13.0*}&\cellcolor{blue!20}\multirow{-2}{*}{-23.6*} &\cellcolor{blue!20} \multirow{-2}{*}{-26.5*}&\cellcolor{blue!20}\multirow{-2}{*}{-26.7} \\
\cline{2-10}
&\multirow{2}{*}{Tib. Ant. (\%)} & \cellcolor{blue!20}& \cellcolor{red!20}&\cellcolor{blue!20} &\cellcolor{red!20}&\cellcolor{blue!20} & \cellcolor{blue!20}&\cellcolor{blue!20} &\cellcolor{blue!20} \\
& &\cellcolor{blue!20} \multirow{-2}{*}{-0.1}&\cellcolor{red!20}\multirow{-2}{*}{6.6*} &\cellcolor{blue!20}\multirow{-2}{*}{-2.7} & \cellcolor{red!20}\multirow{-2}{*}{3.2}&\cellcolor{blue!20} \multirow{-2}{*}{-5.0}& \cellcolor{blue!20}\multirow{-2}{*}{-4.6}&\cellcolor{blue!20}\multirow{-2}{*}{-6.2} &\cellcolor{blue!20}\multirow{-2}{*}{-2.9} \\
\cline{2-10}
&Biological &\cellcolor{blue!20} &\cellcolor{blue!20} &\cellcolor{blue!20} &\cellcolor{blue!20} & \cellcolor{blue!20}&\cellcolor{blue!20} &\cellcolor{blue!20} &\cellcolor{blue!20} \\
&Torque (\%) &\cellcolor{blue!20}\multirow{-2}{*}{-16.7*} &\cellcolor{blue!20}\multirow{-2}{*}{-15.4*} &\cellcolor{blue!20} \multirow{-2}{*}{-14.0*}& \cellcolor{blue!20}\multirow{-2}{*}{-9.9*}&\cellcolor{blue!20}\multirow{-2}{*}{-8.6*} & \cellcolor{blue!20}\multirow{-2}{*}{-14.6*}&\cellcolor{blue!20}\multirow{-2}{*}{-29.6*} &\cellcolor{blue!20}\multirow{-2}{*}{-13.4}\\
\cline{2-10}
&Averaged & & & & & & & & \\
&Peak Assistance (N.m)&\multirow{-2}{*}{22.1} &\multirow{-2}{*}{22.3} &\multirow{-2}{*}{25.5} &\multirow{-2}{*}{18.8} &\multirow{-2}{*}{19.6} &\multirow{-2}{*}{19.6} &\multirow{-2}{*}{21.2} & \multirow{-2}{*}{21.3}\\
 \hline
 \multirow{12}{*}{Subject 3 (30\%)}&\multirow{2}{*}{Gas. Med. (\%)} &\cellcolor{blue!20} &\cellcolor{blue!20} &\cellcolor{blue!20} &\cellcolor{blue!20} &\cellcolor{blue!20} &\cellcolor{blue!20} &\cellcolor{blue!20} &\cellcolor{blue!20}\\
 & &\cellcolor{blue!20}\multirow{-2}{*}{-8.8*} &\cellcolor{blue!20}\multirow{-2}{*}{-1.9} &\cellcolor{blue!20}\multirow{-2}{*}{-8.0*} &\cellcolor{blue!20} \multirow{-2}{*}{-6.8*}& \cellcolor{blue!20}\multirow{-2}{*}{-10.4*}&\cellcolor{blue!20} \multirow{-2}{*}{-11.5*}& \cellcolor{blue!20}\multirow{-2}{*}{-21.2} &\cellcolor{blue!20} \multirow{-2}{*}{-10.4} \\
\cline{2-10}
 &\multirow{2}{*}{Gas. Lat. (\%)} &\cellcolor{red!20} &\cellcolor{blue!20} &\cellcolor{blue!20} &\cellcolor{blue!20} &\cellcolor{blue!20} &\cellcolor{blue!20} &\cellcolor{blue!20} & \cellcolor{blue!20}\\
& & \cellcolor{red!20}\multirow{-2}{*}{20.8*}& \cellcolor{blue!20}\multirow{-2}{*}{-18.7*}&\cellcolor{blue!20}\multirow{-2}{*}{-28.1*} &\cellcolor{blue!20}\multirow{-2}{*}{-20.0*} &\cellcolor{blue!20} \multirow{-2}{*}{-5.9}&\cellcolor{blue!20}\multirow{-2}{*}{-13.2*} &\cellcolor{blue!20}\multirow{-2}{*}{-35.2*} &\cellcolor{blue!20} \multirow{-2}{*}{-20.9}\\
\cline{2-10}
&\multirow{2}{*}{Soleus (\%)} &\cellcolor{blue!20} &\cellcolor{blue!20} &\cellcolor{blue!20} &\cellcolor{blue!20} &\cellcolor{blue!20} &\cellcolor{blue!20} &\cellcolor{blue!20} &\cellcolor{blue!20}\\
& &\cellcolor{blue!20}\multirow{-2}{*}{-21.3*} & \cellcolor{blue!20}\multirow{-2}{*}{-10.0*}&\cellcolor{blue!20}\multirow{-2}{*}{-16.1*} & \cellcolor{blue!20}\multirow{-2}{*}{-20.3*}&\cellcolor{blue!20} \multirow{-2}{*}{-16.9*}&\cellcolor{blue!20}\multirow{-2}{*}{-18.0*} & \cellcolor{blue!20}\multirow{-2}{*}{-27.7*}&\cellcolor{blue!20}\multirow{-2}{*}{-18.9}  \\
\cline{2-10}
&\multirow{2}{*}{Tib. Ant. (\%)} &\cellcolor{red!20} &\cellcolor{red!20} & \cellcolor{blue!20}&\cellcolor{blue!20} &\cellcolor{red!20} & \cellcolor{red!20}&\cellcolor{blue!20} & \cellcolor{blue!20}\\
& &\cellcolor{red!20} \multirow{-2}{*}{7.3*}&\cellcolor{red!20}\multirow{-2}{*}{2.2} &\cellcolor{blue!20}\multirow{-2}{*}{-5.1*} &\cellcolor{blue!20}\multirow{-2}{*}{-9.0*}&\cellcolor{red!20}\multirow{-2}{*}{4.6} &\cellcolor{red!20} \multirow{-2}{*}{0.4}&\cellcolor{blue!20}\multirow{-2}{*}{-18.7} & \cellcolor{blue!20}\multirow{-2}{*}{-4.5}\\
\cline{2-10}
&Biological &\cellcolor{blue!20} &\cellcolor{blue!20} &\cellcolor{blue!20} &\cellcolor{blue!20} &\cellcolor{blue!20} &\cellcolor{blue!20} &\cellcolor{blue!20} &\cellcolor{blue!20}\\
&Torque (\%) &\cellcolor{blue!20}\multirow{-2}{*}{-17.6*} &\cellcolor{blue!20}\multirow{-2}{*}{-15.8*} &\cellcolor{blue!20}\multirow{-2}{*}{-20.8*} &\cellcolor{blue!20}\multirow{-2}{*}{-20.7*} &\cellcolor{blue!20}\multirow{-2}{*}{-17.1*} &\cellcolor{blue!20}\multirow{-2}{*}{-15.0*} &\cellcolor{blue!20}\multirow{-2}{*}{-12.6*} &\cellcolor{blue!20}\multirow{-2}{*}{-19.8} \\
\cline{2-10}
&Averaged & & & & & & & & \\
&Peak Assistance (N.m)& \multirow{-2}{*}{15.2}&\multirow{-2}{*}{16.1} &\multirow{-2}{*}{16.3} &\multirow{-2}{*}{3.3} &\multirow{-2}{*}{13.2} & \multirow{-2}{*}{14.6}&\multirow{-2}{*}{14.5} & \multirow{-2}{*}{14.8}\\
\hline
  \multirow{12}{*}{Subject 4 (40\%)}&\multirow{2}{*}{Gas. Med. (\%)} &\cellcolor{red!20} &\cellcolor{blue!20} &\cellcolor{blue!20}&\cellcolor{blue!20} &\cellcolor{red!20}&\cellcolor{red!20} &\cellcolor{blue!20} &\cellcolor{blue!20}\\
 & &\cellcolor{red!20}\multirow{-2}{*}{0.3} &\cellcolor{blue!20} \multirow{-2}{*}{-2.6}&\cellcolor{blue!20}\multirow{-2}{*}{-1.9} &\cellcolor{blue!20}\multirow{-2}{*}{-8.7*} &\cellcolor{red!20} \multirow{-2}{*}{6.6*}& \cellcolor{red!20}\multirow{-2}{*}{5.4*}&\cellcolor{blue!20} \multirow{-2}{*}{-3.2}&\cellcolor{blue!20}\multirow{-2}{*}{-1.1}\\
 \cline{2-10}
 &\multirow{2}{*}{Gas. Lat. (\%)} &\cellcolor{blue!20} &\cellcolor{red!20} &\cellcolor{blue!20} &\cellcolor{blue!20} &\cellcolor{red!20} &\cellcolor{blue!20} &\cellcolor{red!20} &\cellcolor{blue!20} \\
& &\cellcolor{blue!20}\multirow{-2}{*}{-12.1*} &\cellcolor{red!20}\multirow{-2}{*}{3.5} & \cellcolor{blue!20}\multirow{-2}{*}{-2.1}&\cellcolor{blue!20} \multirow{-2}{*}{-7.3}& \cellcolor{red!20}\multirow{-2}{*}{12.7}&\cellcolor{blue!20}\multirow{-2}{*}{-0.8} & \cellcolor{red!20}\multirow{-2}{*}{11.4}&\cellcolor{blue!20}\multirow{-2}{*}{-1.6}\\
\cline{2-10}
 &\multirow{2}{*}{Soleus (\%)} &\cellcolor{blue!20} &\cellcolor{blue!20} & \cellcolor{blue!20}&\cellcolor{blue!20} &\cellcolor{blue!20} &\cellcolor{blue!20} &\cellcolor{blue!20} &\cellcolor{blue!20} \\
& &\cellcolor{blue!20}\multirow{-2}{*}{-2.8} &\cellcolor{blue!20}\multirow{-2}{*}{-7.1*} &\cellcolor{blue!20} \multirow{-2}{*}{-12.2*}&\cellcolor{blue!20}\multirow{-2}{*}{-5.6*} &\cellcolor{blue!20}\multirow{-2}{*}{-7.0*} &\cellcolor{blue!20}\multirow{-2}{*}{-9.2*} &\cellcolor{blue!20}\multirow{-2}{*}{-7.5} &\cellcolor{blue!20}\multirow{-2}{*}{-8.7}\\
\cline{2-10}
 &\multirow{2}{*}{Tib. Ant. (\%)} &\cellcolor{blue!20} &\cellcolor{red!20} & \cellcolor{blue!20}& \cellcolor{blue!20}&\cellcolor{red!20} &\cellcolor{red!20} & \cellcolor{red!20}& \cellcolor{red!20}\\
& & \cellcolor{blue!20}\multirow{-2}{*}{-0.1}&\cellcolor{red!20}\multirow{-2}{*}{8.2*} & \cellcolor{blue!20}\multirow{-2}{*}{-2.5}&\cellcolor{blue!20} \multirow{-2}{*}{-5.5*}& \cellcolor{red!20}\multirow{-2}{*}{1.1}& \cellcolor{red!20}\multirow{-2}{*}{5.4}&\cellcolor{red!20}\multirow{-2}{*}{3.6} &\cellcolor{red!20}\multirow{-2}{*}{2.4}\\
\cline{2-10}
&Biological &\cellcolor{blue!20} &\cellcolor{blue!20} & \cellcolor{blue!20}&  \cellcolor{blue!20}& \cellcolor{blue!20}& \cellcolor{blue!20}& \cellcolor{blue!20}&\cellcolor{blue!20}\\
& Torque (\%)&\cellcolor{blue!20}\multirow{-2}{*}{-2.7*} &\cellcolor{blue!20}\multirow{-2}{*}{-5.1*} &\cellcolor{blue!20} \multirow{-2}{*}{-7.0*}& \cellcolor{blue!20}\multirow{-2}{*}{-6.0*}&\cellcolor{blue!20}\multirow{-2}{*}{-3.4*} &\cellcolor{blue!20}\multirow{-2}{*}{-3.9*} &\cellcolor{blue!20}\multirow{-2}{*}{-4.7*} &\cellcolor{blue!20}\multirow{-2}{*}{-4.8}\\
\cline{2-10}
&Averaged & & & & & & & & \\
&Peak Assistance (N.m)&\multirow{-2}{*}{22.9} &\multirow{-2}{*}{23.3} &\multirow{-2}{*}{19.7} & \multirow{-2}{*}{17.2}&\multirow{-2}{*}{17.4} &\multirow{-2}{*}{19.1} & \multirow{-2}{*}{19.6}&\multirow{-2}{*}{20.1}\\
\hline
  \multirow{12}{*}{Subject 5 (70\%)}&\multirow{2}{*}{Gas. Med. (\%)} &\cellcolor{blue!20} &\cellcolor{blue!20} &\cellcolor{blue!20}&\cellcolor{blue!20} &\cellcolor{blue!20}&\cellcolor{blue!20} &\cellcolor{red!20} &\cellcolor{blue!20}\\
 & &\cellcolor{blue!20}\multirow{-2}{*}{-7.2*} &\cellcolor{blue!20} \multirow{-2}{*}{-2.5}&\cellcolor{blue!20}\multirow{-2}{*}{-4.5} &\cellcolor{blue!20}\multirow{-2}{*}{-7.9} &\cellcolor{blue!20} \multirow{-2}{*}{-0.9}& \cellcolor{blue!20}\multirow{-2}{*}{-14.6*}&\cellcolor{red!20} \multirow{-2}{*}{7.5}&\cellcolor{blue!20}\multirow{-2}{*}{-45.7}\\
 \cline{2-10}
 &\multirow{2}{*}{Gas. Lat. (\%)} &\cellcolor{blue!20} &\cellcolor{blue!20} &\cellcolor{blue!20} &\cellcolor{blue!20} &\cellcolor{red!20} &\cellcolor{blue!20} &\cellcolor{blue!20} &\cellcolor{blue!20} \\
& &\cellcolor{blue!20}\multirow{-2}{*}{-17.0*} &\cellcolor{blue!20}\multirow{-2}{*}{-19.1*} & \cellcolor{blue!20}\multirow{-2}{*}{-13.4*}&\cellcolor{blue!20} \multirow{-2}{*}{-10.2}& \cellcolor{red!20}\multirow{-2}{*}{22.2*}&\cellcolor{blue!20}\multirow{-2}{*}{-27.2*} & \cellcolor{blue!20}\multirow{-2}{*}{-17.0}&\cellcolor{blue!20}\multirow{-2}{*}{-5.1}\\
\cline{2-10}
 &\multirow{2}{*}{Soleus (\%)} &\cellcolor{blue!20} &\cellcolor{blue!20} & \cellcolor{blue!20}&\cellcolor{blue!20} &\cellcolor{blue!20} &\cellcolor{blue!20} &\cellcolor{blue!20} &\cellcolor{blue!20} \\
& &\cellcolor{blue!20}\multirow{-2}{*}{-23.7*} &\cellcolor{blue!20}\multirow{-2}{*}{-23.9*} &\cellcolor{blue!20} \multirow{-2}{*}{-32.2*}&\cellcolor{blue!20}\multirow{-2}{*}{-21.7*} &\cellcolor{blue!20}\multirow{-2}{*}{-20.3*} &\cellcolor{blue!20}\multirow{-2}{*}{-34.7*} &\cellcolor{blue!20}\multirow{-2}{*}{-17.9*} &\cellcolor{blue!20}\multirow{-2}{*}{-23.1}\\
\cline{2-10}
 &\multirow{2}{*}{Tib. Ant. (\%)} &\cellcolor{blue!20} &\cellcolor{blue!20} & \cellcolor{blue!20}& \cellcolor{blue!20}&\cellcolor{blue!20} &\cellcolor{blue!20} & \cellcolor{blue!20}& \cellcolor{blue!20}\\
& & \cellcolor{blue!20}\multirow{-2}{*}{-35.5*}&\cellcolor{blue!20}\multirow{-2}{*}{-36.6*} & \cellcolor{blue!20}\multirow{-2}{*}{-31.9*}&\cellcolor{blue!20} \multirow{-2}{*}{-23.9*}& \cellcolor{blue!20}\multirow{-2}{*}{-14.4*}& \cellcolor{blue!20}\multirow{-2}{*}{-13.2*}&\cellcolor{blue!20}\multirow{-2}{*}{-34.8*} &\cellcolor{blue!20}\multirow{-2}{*}{-26.0}\\
\cline{2-10}
&Biological &\cellcolor{blue!20} &\cellcolor{blue!20} & \cellcolor{blue!20}&  \cellcolor{blue!20}& \cellcolor{blue!20}& \cellcolor{blue!20}& \cellcolor{blue!20}&\cellcolor{blue!20}\\
& Torque (\%)&\cellcolor{blue!20}\multirow{-2}{*}{-33.7} &\cellcolor{blue!20}\multirow{-2}{*}{-36.4*} &\cellcolor{blue!20} \multirow{-2}{*}{-38.1*}& \cellcolor{blue!20}\multirow{-2}{*}{-36.3*}&\cellcolor{blue!20}\multirow{-2}{*}{-36.9*} &\cellcolor{blue!20}\multirow{-2}{*}{-35.3*} &\cellcolor{blue!20}\multirow{-2}{*}{-33.0} &\cellcolor{blue!20}\multirow{-2}{*}{-35.8}\\
\cline{2-10}
&Averaged & & & & & & & & \\
&Peak Assistance (N.m)&\multirow{-2}{*}{34.7} &\multirow{-2}{*}{34.2} &\multirow{-2}{*}{38.7} & \multirow{-2}{*}{29.6}&\multirow{-2}{*}{31.0} &\multirow{-2}{*}{30.1} & \multirow{-2}{*}{34.5}&\multirow{-2}{*}{33.8}\\

\hline
  \multirow{12}{*}{Subject 6 (40\%)}&\multirow{2}{*}{Gas. Med. (\%)} &\cellcolor{red!20} &\cellcolor{red!20} &\cellcolor{red!20}&\cellcolor{red!20} &\cellcolor{red!20}&\cellcolor{red!20} &\cellcolor{red!20} &\cellcolor{red!20}\\
 & &\cellcolor{red!20}  \multirow{-2}{*}{26.8*}&
\cellcolor{red!20}  \multirow{-2}{*}{1.9}&
\cellcolor{red!20}  \multirow{-2}{*}{14.1*}&
\cellcolor{red!20}  \multirow{-2}{*}{26.4*}&
\cellcolor{red!20}  \multirow{-2}{*}{16.1*}&
\cellcolor{red!20}  \multirow{-2}{*}{19.2*}&
\cellcolor{red!20}  \multirow{-2}{*}{2.8}&
\cellcolor{red!20}  \multirow{-2}{*}{20.2}\\
 \cline{2-10}
 &\multirow{2}{*}{Gas. Lat. (\%)} &\cellcolor{blue!20} &\cellcolor{blue!20} &\cellcolor{blue!20} &\cellcolor{blue!20} &\cellcolor{blue!20} &\cellcolor{blue!20} &\cellcolor{blue!20} &\cellcolor{blue!20} \\
& &
\cellcolor{blue!20}  \multirow{-2}{*}{-12.3*}&
\cellcolor{blue!20}  \multirow{-2}{*}{-25.4*}&
\cellcolor{blue!20}  \multirow{-2}{*}{-24.7*}&
\cellcolor{blue!20}  \multirow{-2}{*}{-69.7*}&
\cellcolor{blue!20}  \multirow{-2}{*}{-73.8*}&
\cellcolor{blue!20}  \multirow{-2}{*}{-24.6*}&
\cellcolor{blue!20}  \multirow{-2}{*}{-61.3*}&
\cellcolor{blue!20}  \multirow{-2}{*}{-54.7}\\
\cline{2-10}
 &\multirow{2}{*}{Soleus (\%)} &\cellcolor{blue!20} &\cellcolor{blue!20} & \cellcolor{blue!20}&\cellcolor{blue!20} &\cellcolor{blue!20} &\cellcolor{blue!20} &\cellcolor{blue!20} &\cellcolor{blue!20} \\
& &
\cellcolor{blue!20}  \multirow{-2}{*}{-23.2*}&
\cellcolor{blue!20}  \multirow{-2}{*}{-20.8*}&
\cellcolor{blue!20}  \multirow{-2}{*}{-8.5*}&
\cellcolor{blue!20}  \multirow{-2}{*}{-25.2*}&
\cellcolor{blue!20}  \multirow{-2}{*}{-22.2*}&
\cellcolor{blue!20}  \multirow{-2}{*}{-20.9*}&
\cellcolor{blue!20}  \multirow{-2}{*}{-28.4*}&
\cellcolor{blue!20}  \multirow{-2}{*}{-19.7}\\
\cline{2-10}
 &\multirow{2}{*}{Tib. Ant. (\%)} &\cellcolor{red!20} &\cellcolor{red!20} & \cellcolor{red!20}& \cellcolor{red!20}&\cellcolor{red!20} &\cellcolor{red!20} & \cellcolor{red!20}& \cellcolor{red!20}\\
& & 
\cellcolor{red!20} \multirow{-2}{*}{9.1*}&
\cellcolor{red!20} \multirow{-2}{*}{0.6}&
\cellcolor{red!20}  \multirow{-2}{*}{31.2*}&
\cellcolor{red!20}  \multirow{-2}{*}{5.3}&
\cellcolor{red!20}  \multirow{-2}{*}{25.4*}&
\cellcolor{red!20}  \multirow{-2}{*}{56.6*}&
\cellcolor{red!20}  \multirow{-2}{*}{11.7} &
\cellcolor{red!20}  \multirow{-2}{*}{21.7}\\
\cline{2-10}
&Biological &\cellcolor{blue!20} &\cellcolor{blue!20} & \cellcolor{blue!20}&  \cellcolor{blue!20}& \cellcolor{blue!20}& \cellcolor{blue!20}& \cellcolor{blue!20}&\cellcolor{blue!20}\\
& Torque (\%)&
\cellcolor{blue!20}  \multirow{-2}{*}{-19.3*}&
\cellcolor{blue!20}  \multirow{-2}{*}{-22.1*}&
\cellcolor{blue!20}  \multirow{-2}{*}{-13.1*}&
\cellcolor{blue!20}  \multirow{-2}{*}{-27.8*}&
\cellcolor{blue!20}  \multirow{-2}{*}{-29.8*}&
\cellcolor{blue!20}  \multirow{-2}{*}{-19.9*}&
\cellcolor{blue!20}\multirow{-2}{*}{-29.3*}
 &\cellcolor{blue!20}\multirow{-2}{*}{-21.0}\\
\cline{2-10}

&Averaged & & & & & & & & \\
&Peak Assistance (N.m) &
 \multirow{-2}{*}{28.8}&
 \multirow{-2}{*}{29.3}&
 \multirow{-2}{*}{32.4}&
 \multirow{-2}{*}{23.3}&
 \multirow{-2}{*}{23.3}&
 \multirow{-2}{*}{24.3}&
\multirow{-2}{*}{24.6}&\multirow{-2}{*}{26.8}
\\
\hline

\end{tabular}
\footnotesize
  \begin{tablenotes}
   \item[\tnote{*}] statistical significance (P \textless 0.007)
  \end{tablenotes}
 \end{threeparttable}
\end{center}
\end{table*} 

\section{Results}

Fig. \ref{fig_all} presents the different transformation realized in the NMBC from EMG (Fig. \ref{fig_all}-A) and joint position (Fig. \ref{fig_all}-B) to muscle forces (Fig. \ref{fig_all}-C) and joint torques (Fig. \ref{fig_all}-D) for the walking tasks of 0.5 m/s and -5\% elevation. In this figure, the results are presented as an average over the gait cycle. 
In Fig. \ref{fig_all}-D, the biological joint torque for the two conditions as well as the exoskeleton torque are presented over the gait cycle. It can be observed that a reduction of biological joint torque is always achieved for the full gait cycle when assisted for the 0.5 m/s, -5\% elevation walking task. Similar results can be observed for the rest of the tested walking tasks (see appendix \ref{ann::fig} Fig. S\ref{fig_ik}-A).
In Fig. \ref{fig_all}-A, the EMGs over the gait cycle for all the recorded muscles during the 0.5 m/s, -5\% elevation walking task are presented. The calf muscles present the most reduction during the push-off phase, for the Tibialis Anterior most of the reduction happens during the start of the swing phase.
The following sub-sections are presenting averaged results and reduction percentages.

\subsection{Biological torque reduction}
Fig. \ref{fig_adpatation} shows the biological ankle torque averaged across each gait cycle (for all tested walking conditions) for the two tested exoskeleton conditions (\textit{i.e.} minimal impedance and assistance from NMBC) as well as the trends (first-order polynomial fitting curve) (dashed line) for the tested walking conditions (red) and during the transition (green). 
Fig. \ref{fig_torque_bar} presents the estimated biological ankle torque for all tested walking conditions as well as for all transitions and for the complete experiment between the two exoskeleton conditions. Reductions between exoskeleton conditions were achieved for biological ankle torque for all tested walking conditions (ranging from 20\% to 24\%) and reduction of 24\% was obtained also during the transition between walking conditions. The final reduction of all participants for the complete experiment (all tested walking tasks and transition between those tasks) was 22\%.
Fig. \ref{fig_torque_desnity} shows the distribution of the biological torques for the two tested exoskeleton conditions for all walking condition and all subjects. Significant reductions were found for all walking conditions as well as walking condition transition, and that for all subjects.

Table \ref{table:1} presents ankle torque reduction between exoskeleton conditions obtained for each subject independently. It can be seen that for all subjects and all walking conditions, the reduction in biological torque was always positive (blue cell color) (maximum 38.1\% (subject 5 for 0.7 m/s, 12\%) minimum 3.4\% (subject 4 for 0.5 m/s, 0\%)). Table \ref{table:1} also shows the averaged peak assistance in N.m. received by subjects and the individual support ratio. 
The moonwalking task showed a significant reduction in biological joint torque of 40\%. (Fig. \ref{fig_EMG_moonwalk})

\subsection{EMG reduction}
Fig. \ref{fig_EMG} shows EMG reduction for of all muscles across all walking conditions (between 15\% and 6\%) as well as during the transition between walking conditions (14\%) and for the complete experiment (12\%). Fig. \ref{fig_EMG} presents detailed results at the muscle level. It can be observed that reduction was obtained for all calf muscles and all walking conditions, during transitions as well as for the complete experiment (maximum reduction of 32.8\% for the Gastrocnemius lateralis during transition). 

Table \ref{table:1} presents the EMG reduction obtained for each subject independently. It can be observed that the Soleus muscle always presented reduction in EMG (blue cell color) for all subjects (maximum 40.6\% (subject 1 for 2.8 km/h, -5\%) and minimum 2.8\% (subject 4 for 2.8 km/h, -5\%)). The Tibialis Anterior had the least EMG reduction across all muscles with an increased in amplitude for subject 1 and 6 during assisted exoskeleton condition when compared to non-assisted exoskeleton condition. However, EMG increase was observed only during the swing phase where overall muscle activation and resulting torque are the lowest. Averaged EMG over gait cycle can be seen in Fig. S\ref{fig_EMG_sup}-A.

The moonwalking task showed reduction in EMG (Fig. \ref{fig_EMG_moonwalk}) for all muscles (from maximum 30\% (Gastrocnemius Medialis) to 17\% (Soleus)) and significance was found for all muscles except the Tibialis anterior.

\subsection{Human-exoskeleton torque invariance across exoskeleton conditions}
The total human+exoskeleton torque between exoskeleton conditions (summation of the estimated biological ankle joint torque and exoskeleton delivered torque) was preserved across walking conditions (between -2\% for 0.7 m/s at -5\% inclination and 0.5 m/s at 0\% inclination and 6\% for 0.5 m/s at 12\% inclination). The total human-exoskeleton torque between exoskeleton conditions was also preserved for the complete experiment (2\%).


\section{Discussion}
%
NMBC's ability of decoding biological ankle joint torques from multiple wearable EMGs and joint angle sensors, enabled six different subjects to voluntarily control bi-lateral ankle exoskeletons across six walking conditions and all relative transitions over more than 18 minutes of continuous walking experiment (see supplementary movie 1). Across all subjects, walking conditions and transitions, the NMBC-controlled exoskeleton systematically reduced biological joint torques and EMGs when compared to non-assisted walking (Figs \ref{fig_torque_bar}, \ref{fig_torque_desnity}, \ref{fig_all} and \ref{fig_EMG}).
In a case-study, NMBC allowed one subject to control the bilateral ankle exoskeleton during moonwalking while at the same time reducing the subject's muscular effort for all recorded muscles (Fig. \ref{fig_EMG_moonwalk}). This is in need of further validation on a wide group of subjects to further show the extrapolation capacity of the controller.

State of the art exoskeleton control approaches are based on
pre-defined torque profiles or state machines \cite{asbeck2015biologically, Galle2017ReducingPower, Kim2019ReducingExosuit, Ding2018Human-in-the-loopWalking, Khazoom2019DesignLanding, Zhang2017HumanInTheLoop}. Although real-time mechanistic models of neuromuscular reflexes were proposed to control prostheses and exoskeletons, these were not driven by \textit{in vivo} biomechanical data, but by a finite set of \textit{a priori} chosen reflexive rules \cite{Geyer2010AActivities, Tamburella2020NeuromuscularSubjects} (\textit{e.g.} positive force feedback and stretch reflexes). Moreover, existing human-in-the-loop optimization techniques require several tens of minutes (up to 45 minutes) for the generation of appropriate exoskeleton torque commands \cite{Zhang2017HumanInTheLoop}, hampering support of large repertoire of movements. As a result, existing approaches are limited to pre-defined movements and do not enable voluntary control of exoskeletons.

Contrary to a proportional EMG controller, NMBC realizes a non-linear bio-inspired sensor fusion between multiple kinematic inputs (joint angles) and bio-electrical input signals (EMGs) through a muscle model that filters out the high level frequency components of the EMG (spring and damper contained in the muscle model). The effective cut-off frequency of such "muscle-inspired filter" is dynamically modulated as a function of joint kinematics, dictating force-length-velocities dependencies. This prevents transferring high-frequency EMG artefacts along the modelling and control pipeline, which would otherwise result in non-physiological torque output (Fig. \ref{fig_Hill}). Moreover, the musculoskeletal geometry model provides biomechanically consistent moment arms. These acted as non-linear weighting coefficients for the projection of EMG-dependent muscle force to the joint. As opposed to state of the art EMG direct controllers \cite{McCain2019MechanicsControl}, NMBC enables fusing higher-dimensional EMG signals onto a lower-dimensional set of joint torques.%


In contrast to the state of the art, the proposed NMBC used an optimization-based calibration conducted once per subject. Results showed that on average for each individual, the calibration procedure required kinematic and kinetic data relative to two minutes of ground-level walking without exoskeleton. 
After calibration, NMBC transformed EMG signals to decode resulting muscle and joint torques underlying any movement condition with no assumption on what neuromuscular reflexes to be modelled. When used for exoskeleton control, NMBC reduced torque and EMGs in unseen walking conditions (\textit{i.e.} not used during the calibration procedure) without having to change low-level control parameters or having to switch across pre-defined states or motor torque profiles. Moreover, since no exoskeleton was worn during the walking trials for calibration, current results showed NMBC's ability to account for the exoskeleton added loading during real-time control tests. This all shows evidence of NMBC's ability of extrapolating across walking conditions as well as different load cases. 

Our results can be compared with EMG reductions realized with other alternative methods, e.g. \cite{Zhang2017HumanInTheLoop} using human in the loop optimization, reported a reduction of 41\% for the Soleus with respect to zero torque condition (similar to our non-assisted condition) at a walking speed of 1.25 m/s. In \cite{Collins2015ReducingExoskeleton}, using a passive exoskeleton, the Soleus was reduced by 22\% and biological ankle torque by 14\% at a walking speed of 1.25 m/s. In \cite{Galle2017ReducingPower}, using a pre-computed torque pattern, the authors obtained a reduction of Soleus of 30\%, 20\% for the Gastrocnemius Medialis and an increase of 100\% for the Tibialis Anterior at a walking speed of 1.25 m/s. Finally, in \cite{jackson2019heuristic}, the authors showed a reduction of Soleus EMG of 32\% and an increase of Tibialis Anterior (\% not specified) at a walking speed of 1.25 m/s using the human in the loop method with EMG level reduction as the objective function. In comparison, we obtained a reduction of 20.5\% for the Soleus, 6.57\% for the Gastrocnemius Medialis, a decrease of 0.36\% for the Tibialis Anterior and a decrease of 22\% for the biological joint torque. 
The results from the NMBC are on par with the current state of the art in exoskeleton mediated muscle effort reduction with the main difference being that our results encompass a multitude of walking conditions and the transitions between those which is currently not possible using a passive exoskeleton system or active exoskeleton using pre-computed torques profiles.

In this study we did not consider a no-exo condition as the weight of the device was substantial, i.e. 10 kg for the backpack and 5 kg per side. Nevertheless, from \cite{Zhang2017HumanInTheLoop} the difference in Soleus reduction between the no-exo and zero torques condition represents only a 5\% difference. Our results showed an average of 20\% reduction in the soleus’ EMG (Complete experiment), which would correspond to a 15\% reduction for a possible no-exo conduction if the same exoskeleton as Zhang \textit{et al} \cite{Zhang2017HumanInTheLoop} would have been used.

Results showed NMBC's ability of dynamically adapting to the mechanical demand of each walking condition. Fig. \ref{fig_adpatation} and Fig S\ref{fig_ik}-A from the appendix \ref{ann::fig} show that NMBC prescribed more torque to the exoskeleton during more mechanically demanding walking conditions (\textit{i.e.} higher speeds and ground elevations), with the proportion of human-contributed joint torque proportionally decreasing. This provides evidence of NMBC's ability of responding to different mechanical demands as required in real-world environments. 
%
Users' kinematics between conditions were similar with the main difference being at the level of the ankle joint angle that showed a reduced plantar flexion during push off (see appendix \ref{ann::fig} Fig. S\ref{fig_ik}-B). This can be explained by the added joint torque given by the assistance and reduced muscle force needed to be produced by the user (see Fig. \ref{fig_all}, Fig. S\ref{fig_ik}-A and S\ref{fig_EMG_sup}-B in appendix \ref{ann::fig}). Since the muscles do not have to produce as much force compared to the non assisted condition, they do not need to reach a higher force production, thus working at shorter length in the force length relationship of the muscle (Fig. \ref{fig_Hill}).

Within each condition, the total amount of joint torque generated by the human+exoskeleton system during walking (human-generated + exoskeleton-generated torque) was always preserved between assisted and non-assisted walking (Fig. \ref{fig_torque_bar}). This provided indirect validation of NMBC-estimated biological torques, \textit{i.e.} for each individual, walking speed and foot strike cadence were controlled and therefore were preserved across assisted and non-assisted conditions. As a results, the net human+exoskeleton ankle joint torque was expected to be similar across assisted and non-assisted walking, as shown in our results. Moreover, this showed that, through NMBC, the human and the exoskeleton were always capable of converging towards an equilibrium, during which human walking was more economical in terms of EMG and biological torque generation. This is crucial to promote user's acceptance towards wearable assistive robots. 

Remarkably, since even small discrepancies in the onset time between the biological joint rotation and parallel exoskeleton motor actuation could potentially increase biological EMGs and joint torques during walking \cite{Zhang2017HumanInTheLoop}, our results demonstrate that NMBC could precisely synchronize the exoskeleton actuation with human muscle contraction. Timely torque delivery to biological joints has been shown to be crucial for metabolic energy reduction notably during the push-off phase \cite{caputo2014universal}. Metabolic energy comparison via a respiratory system will be conducted in future work to compare net metabolic reduction. Future work will also test NMBC to control lightweight exoskeletons in out-of-the-lab scenarios.

Results showed that the Tibialis anterior muscle underwent the least EMG reduction and even a constant increase for subject 1 and 6. This increase of the EMG for the tibialis anterior could be explained by the fact that the average joint torque and assistance for all tasks (see Fig. S\ref{fig_sbj1} in supplementary figure) for subject 1 presents very little or no dorsiflexion torque and provided very little or no dorsiflexion assistance. In this case, the controller was not providing assistance, potentially even counter-assisting (i.e. in the plantar flexion direction) due to possible underestimation of dorsiflexion torque.


The authors previously demonstrated that real-time model-based controllers enabled post-stroke and incomplete spinal cord injury subjects to control a uni-lateral robotic exoskeleton to perform knee and ankle joint rotations executed from seated positions \cite{durandau2019voluntary}. This provided evidence that data-driven model-based control strategies have potentials to be translated and personalized to individuals with neuro-musculo-skeletal injuries. Authors also employed model-based controllers to enable a transradial amputee to control a uni-lateral wrist-hand prosthesis \cite{sartori2018robust} as well as healthy individuals to control a uni-lateral elbow joint soft exosuit \cite{8963852}. 
With respect to our previous work, the current paper shows for the first time that the combination of calibrated person-specific data-driven neuromechanical modelling and disturbance observers can enable stable yet voluntary control of a complex bi-lateral exoskeleton. This, was observed during complex movement scenarios, \textit{i.e.} a broad range of walking conditions and transitions underlying the coordination of a large number of muscles as well as during the exchange of large interaction forces between the ground, the exoskeleton and the human body, something not achieved previously. In this context, combination of calibration, modelling and a passive disturbance observer was crucial to assure stable exoskeleton operation in response to large human-exoskeleton-ground interaction forces. This all is crucial for enabling robotic exoskeleton applications outside of the lab in unseen and unstructured terrains. 

This study was affected by a number of limitations. 
The low number of subjects (6) is a limitation of this experiment. Nevertheless, the systematic reduction in muscle effort for all subjects gives confidence on the validity of the methods.

Different support ratios were identified to each subject based on feedback from the subjects to offer them comfort of use. Selection of assistance level by the user are not always based on the maximisation of metabolic reduction but on comfort, pain, stability which are more difficult to quantified and are different between subjects \cite{Ingraham2020UserExoskeletons}. The results showed that for subjects 2, 4 and 6, for the same level of support ratio different levels of reduction were found (see Table 1). These different results may be explained by the fact that different subjects (having different exoskeleton experience levels) react to the exoskeleton assistance in a different way, yielding differences in EMG/torque reduction even if the same support ratio is used. 
Other aspects could also explain subject-specific differences in support ratios such as the quality of the model calibration, the accuracy of the maximal voluntary contraction tasks for EMG normalization, the EMG sensors placements across the calibration session day and the actual experimental session day.

The relation between support ratio and muscle effort reduction was not fully explored in this study and is one of the current limitations. What can be extracted for the current results is that, to a certain limit, an increase in the support-ratio could increase the reduction in joint torque as shown between subjects 1 and 3 but reduction between subjects can vary even for the same support ratio (see subjects 2 and 4).
Imposing the same support ratio across all subjects would have allowed for a more controlled comparison. However, it would have not been practically possible due to the above mentioned aspects. This is a limitation of our study. Future work will integrate subject training to control for individual subject response to exoskeleton assistance. In this context, \cite{Poggensee2021HowAssistance} previously showed that exoskeleton training may contribute up to 50\% of the metabolic reduction, with 4 hours of training being needed to observe full benefits. Another explanation could be the inter-subject physiological variability, the difference in forces or muscle-tendon stiffness could allow for an acceptance to a higher level of support.
With this in consideration, a more systematic study on the effect of support gains should be conducted.


Future work will devise automatic selection of support ratios based on individual neuromuscular function. Moreover, support ratios were kept constant throughout the experiment. This led to small unwanted ankle joint rotations being occasionally induced by the exoskeleton during the swing phase, i.e. when ankle joint stiffness was lowest. Future work, will address this point by dynamically modulating the exoskeleton support ratio based on joint stiffness estimates, to provide more assistance when the joint is stiffer (\textit{i.e.} push-off) and less assistance when the joint is lax (\textit{i.e.} swing) \cite{sartori2015modeling,cop2019model}.

Further limitation of our controller is that the muscle model does not take into account fatigue and the tests realised were done on healthy subjects that should not get fatigue during our relatively short experiment (\textless 20 min). Estimation of muscle fatigue and the change of muscle parameters over time (due to training for example) are research questions that we hope to explore in the future as they could have a direct effect on the efficiency of assistance delivered by exoskeletons over a long period.
The real-time neuromusculoskeletal model was previously validated but without the use of an exoskeleton \cite{durandau2017robust}. The estimated joint torque of the exoskeleton user was not validated and its accuracy was not quantified.
Validation of the estimated joint torque under assistance and with the added exoskeleton needs to be carried on in the future but limitation on the modelisation of the interaction force between human and exoskeleton needs to be solved first. Otherwise, the joint torque validation using inverse dynamics would not result in meaningful and trustworthy results. It is noteworthy that the exoskeleton assistive torque reduced muscular effort in unseen tasks, which would not be possible if the framework was not predicting biomechanically plausible joint torques and in such systematic ways i.e. reduction of EMG and joint torque for all subjects.

During assistance, reduction in joint ankle angle can be observed during push-off (fig 3). This has already been seen in similar studies \cite{Kim2019ReducingExosuit,Collins2015ReducingExoskeleton}. In future work, we would like to better understand how provided assistance by exoskeleton changes the kinematics and dynamics of the human joint and muscle-tendon system and what can be the advantages and disadvantages of such changes.

Another limitation is the practicality of the method. The two main issues for the broad adoption of this method are the use of single surface EMG sensors and the calibration procedure. Surface EMGs are sensitive to placements and maximal voluntary contraction for its normalization. Those can easily vary depending on the expertise of the person placing these sensors. For this, the use of sensorised stretchable textile-based garments for recording high-density EMGs non-obstructively \cite{Farina2010High-densityProstheses} in combination with blindsource
separation techniques will enable establishing a direct connection between the robotic exoskeleton and human spinal motor neurons \cite{Farina2016Man-MachineReinnervation,chen2020adaptive}. This has the potential to lead to an NMBC that is more intuitive and less sensitive to electrode placement and signal normalization as previously discussed \cite{ Sartori2017InVivo}.
For the calibration, the tasks needed for the calibration dataset can sometimes be challenging for patients. For example, spinal cord injury patients that are wheelchair-bound cannot walk on a force plate (ground reaction forces are needed for inverse dynamics). A workaround would be to get informed directly by muscle parameters using imaging techniques (ultrasound and MRI) to better constrain the muscle model parameters using, for example, experimental recorded tendon length, fiber length, pennation angle and physiological cross-section area. 
For healthy users, removing the calibration session (day one of our experimental protocol) would increase the usability of the system. This could be done by calibrating the model online using exoskeleton sensor data while the user is walking with the exoskeleton in transparent mode. This model calibration needs to be done only once or when a structural change in the musculoskeletal system (increase or decrease of muscle mass) happens.

\section{Conclusion}
We presented a new human-exoskeleton interface that combined subject-specific data-driven neuromechanical modelling with low-level disturbance observers within a real-time control framework termed NMBC. The study confirmed the two research questions presented in the introduction. First, significant biological joint torque reduction was obtained across all walking conditions and subjects. Second, EMG reduction was obtained for the Soleus muscles on all conditions and all subjects. These results showed that the NMBC enabled individuals to control a robotic bi-lateral exoskeleton voluntarily during a broad range of locomotion conditions and transitions as well as dexterous and challenging task (moonwalking, for one subject). Moreover, NMBC enabled exoskeleton dynamic adaptation to motor tasks mechanical demands over unseen locomotion conditions that were not considered for the NMBC calibration stage and with no need to use pre-defined locomotion mode classification or state machines. This represents an important step to enable the use of wearable robots outside of the lab to support complex movements during real-life situations.

%
\appendices

\section{Muscle activation computation} \label{ann::muscleActivation}

Activation is computed using the following equation to capture the non-linear twitch response of muscle fibers:
\begin{equation}
A(t)=\frac{(e^{E\overline{u}(t)}-1)}{(e^{E}-1)}
\end{equation}
With $E$ being the EMG shape factor and $\overline{u}(t)$ being the filtered and normalized EMG.

\section{Muscle-tendon force computation} \label{ann::MTU}

Muscle-tendon force is computed using the following equation:
\begin{equation}
F^{MTU}(t)=F^{T}(t)=F^{M}(t)cos(\alpha(t))
\end{equation}With $F^{MT}$ representing the muscle-tendon force, $F^{T}$ is the tendon force, $F^{M}$ is the muscle fiber force and $\alpha$ is the muscle fibers pennation angle.
Muscle fiber force is computed using the following equation:
\begin{equation}
\begin{split}
F^{M}(t)=F^{Max}_{Iso}(\overline{f_{a,l}}(\overline{L^{M}(t)})\overline{f_{a,v}}(\overline{V^{M}(t))A}(t)+ \\
\overline{f_{p}}(\overline{L^{M}(t)})+D^{M}\overline{V^{M}(t)})
\end{split}
\end{equation}

With $F^{Max}_{Iso}$ representing the maximal isometric muscle force, $\overline{f_{a,l}}(\overline{L^{M}})$ is the active normalized muscle force-length relationship (Fig. 2-A), $\overline{f_{a,v}}(\overline{V^{M}})$ is the active normalized muscle force-velocity relationship (Fig. 2-B), $A(t)$ is the muscle activation from eq. 1, $\overline{f_{p}}(\overline{L^{M}})$  is the passive normalized muscle force-length relationship (Fig. \ref{fig_Hill}-A), $D^{M}$ is a linear damping coefficient, $\overline{L^{M}}$  is the muscle length normalized by $L^{M}_{Opt}$, $L^{M}_{Opt}$ the optimal fiber length representing the muscle length at which the muscle produce the maximal force $F^{Max}_{Iso}$ and $\overline{V^{M}}$ is the muscle velocity normalized by $10 * L^{M}_{Opt}  m/s$.

\section{Tendon force computation}\label{ann::MT}

The tendon strain is obtained using the following equation:
\begin{equation}
S^{T}(t)=\frac{L^{T}(t)-L^{T}_{Slack}}{L^{T}_{Slack}}
\end{equation}
With $S^{T}$ the tendon strain,  $L^{T}$ is the tendon length and $L^{T}_{Slack}$ is the tendon slack length, which is the length beyond which the tendon starts generating resistive force.
As dynamic contractions occur in the muscle, change in fiber kinematics results in pennation angle $\alpha$ change, while the overall muscle thickness is assumed constant in the presented model (nominator in eq.5). Pennation angle $\alpha$ is continuously updated at each time instant $t$ using the following equation:
\begin{equation}
\alpha(t)=\arcsin(\frac{L^{M}_{Opt}\sin(\alpha_{Opt})}{L^{M}})
\end{equation}
With  $\alpha_{Opt}$ being the pennation angle when muscle fibers are at $L^M_{Opt}$.

\section{Moment arm computation} \label{ann::MA}

The moment arm is computed with the following equation:
\begin{equation}
r = \frac{d L^{MT}}{d\theta}
\end{equation}
with r the moment arm, $L^{MT}$ is the muscle-tendon length and $\theta$ is the joint angle.

\section{Pre-tuning of optimal fiber length and tendon slack length} \label{ann::pre-calib}

The pre-tuning procedure \cite{winby2008evaluation} is done using a non-linear optimization procedure (interior point optimizer \cite{wachter2006implementation}) that minimizes the difference between muscle-tendon length determined from a scaled geometry model and predicted muscle-tendon length across a range of nominal joint angles as reported below:
\begin{equation}
min\sum^{11}_{i=1}(L^{MT}(i)-L^{MT}_{Pred}(i))^2
\end{equation}
With $L^{MT}(i)$ the muscle tendon length from the scaled model, $i$ represents the index for eleven angles equally spaced across the full range of motion of the DOFs crossed by the considered muscle and $L^{MT}_{Pred}(i)$ the predicted muscle tendon length computed using the following equation:
\begin{equation}
\begin{split}
L^{MT}_{Pred}(i)=L^{T}_{Slack} + L^{T}_{Slack}\varepsilon_{T}(i)+L^{M}_{Opt} \\
\overline{L^{M}}(i)\cos{(\alpha(i))}
\end{split}
\end{equation}
With $\overline{L^{M}}(i)$ the normalized muscle length determined from the unscaled model (\textit{i.e.} using cadaveric study data) and $\varepsilon_{T}(i)$ computed using the following equation:
\begin{equation}
\begin{split}
\varepsilon_{T}=\frac{\overline{F^{m}}\cos(\alpha) + 0.2375}{37.5}\text{ for }\varepsilon_{T}>0.0127 \\
    \varepsilon_{T}=\frac{\ln({\frac{\overline{F^{m}}\cos{\alpha}}{0.06142}+1})}{124.929}\text{ for } \varepsilon_{T}\leq0.0127
\end{split}
\end{equation}
With $\overline{F^{m}}$ the normalized muscle force during maximum muscle activation from the unscaled model.

The values for Eq.9 were taken directly from the references \cite{winby2008evaluation}. Where 0.0127 represents a cut-off (1.27\%) when the tendon strain is greater than 1.27\%, tendon force can be computed as a linear function by
\begin{equation}
 \bar{F^T}=37.5\varepsilon^t-0.2375 
 \end{equation}
see \cite{Zajac1989MuscleControl., Manal2004Subject-specificMethod} with
\begin{equation}
\bar{F^T}=\bar{F^M}cos(\alpha).
 \end{equation}
When the tendon strain is smaller than 1.27\% (or also call the toe region) it can be computed by an exponential function:
\begin{equation}
\bar{F^T}=0.06142e^{124.929\varepsilon^t}-1
 \end{equation}

\section{Calibration objective function} \label{ann::calib}
The objective function for the calibration is the following:
\begin{equation}
\frac{1}{N_{Rows}N_{Trials}N_{DOFs}}\sum^{N_{Trials}}\sum^{N_{DOFs}}\sum^{Rows}(\tau_{Pred}-\tau_{ID})^2
\end{equation}
With $N_{Rows}$ the number of data points for the considered trials, $N_{DOFs}$ is the number of degree of freedoms, $N_{Trials}$ is the number of trials, $\tau_{Pred}$ the joint torque computed by our model presented in the previous section and $\tau_{ID}$ the experimental joint torque.
 
\section{Muscle force relationships splines coefficient} \label{ann::MFRCoeff}
The muscle force relationships B-splines are based on the following data (see \cite{Zajac1989MuscleControl.}):\\
Active Force Length relationship:\\
      xPoints: -5 0 0.401 0.402 0.4035 0.52725 0.62875 0.71875 0.86125 1.045 1.2175 1.43875 1.61875 1.62 1.621 2.2 5
      \\ yPoints: 0 0 0 0 0 0.226667 0.636667 0.856667 0.95 0.993333 0.77 0.246667 0 0 0 0 0
      
Passive Force Length relationship:
      \\ xPoints: -5 0.998 0.999 1 1.1 1.2 1.3 1.4 1.5 1.6 1.601 1.602 5
     \\ yPoints: 0 0 0 0 0.035 0.12 0.26 0.55 1.17 2 2 2 2
     
ForceVelocity relationship:
      \\  xPoints: -10 -1 -0.6 -0.3 -0.1 0 0.1 0.3 0.6 0.8 10
      \\  yPoints: 0 0 0.08 0.2 0.55 1 1.4 1.6 1.7 1.75 1.75 “

\setcounter{figure}{0} 
\renewcommand{\figurename}{Figure S}
\section{Supplementary figure} \label{ann::fig}
\begin{figure}[h]
\centering
\includegraphics[width=0.5\textwidth]{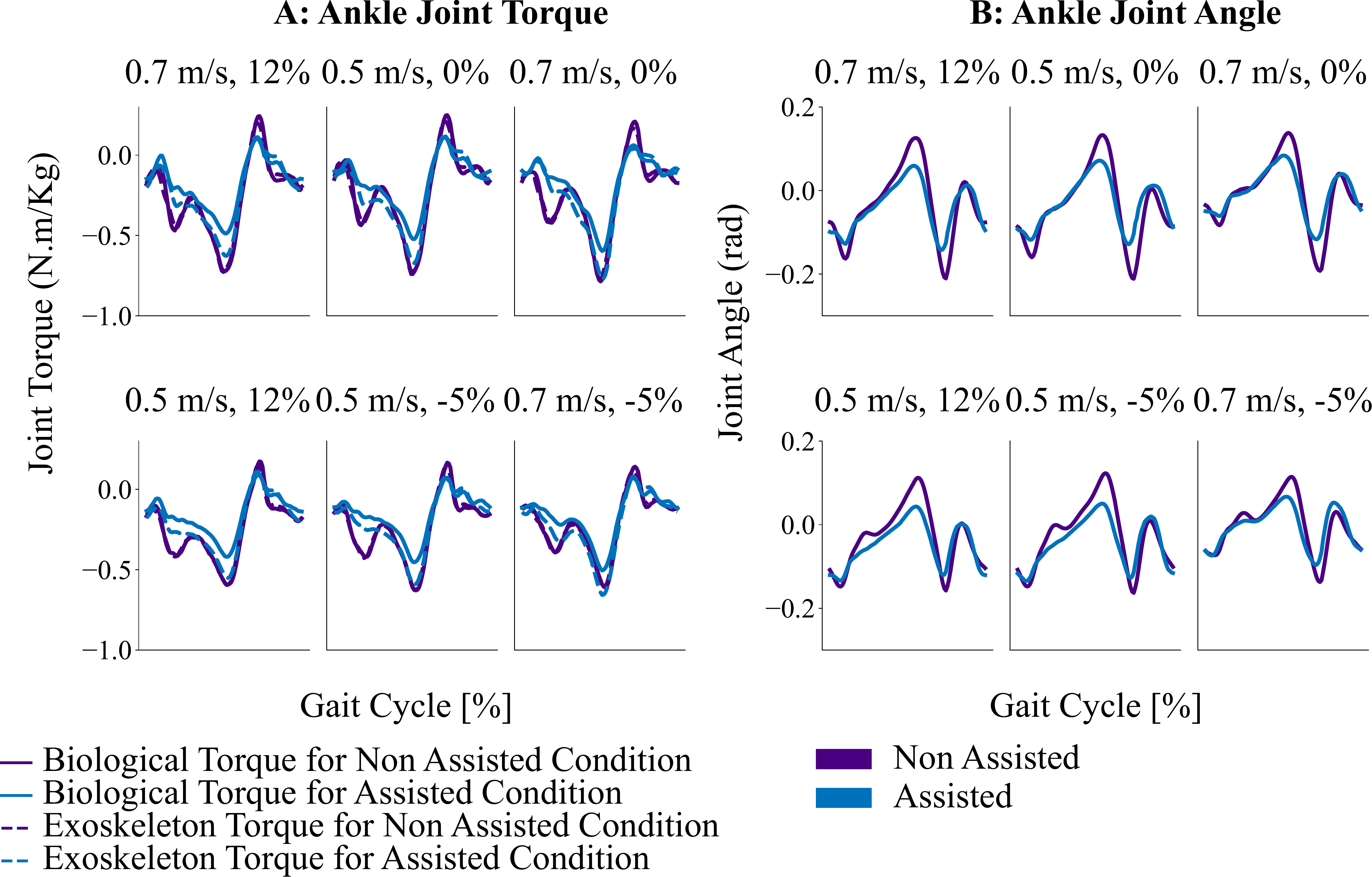}
\caption{Averaged joint angles (A) and joint torques (B) of the ankle for all subjects and for the two tested conditions (assisted and non-assisted).}
\label{fig_ik}
\end{figure}



\begin{figure}[h]
\centering
\includegraphics[width=0.5\textwidth]{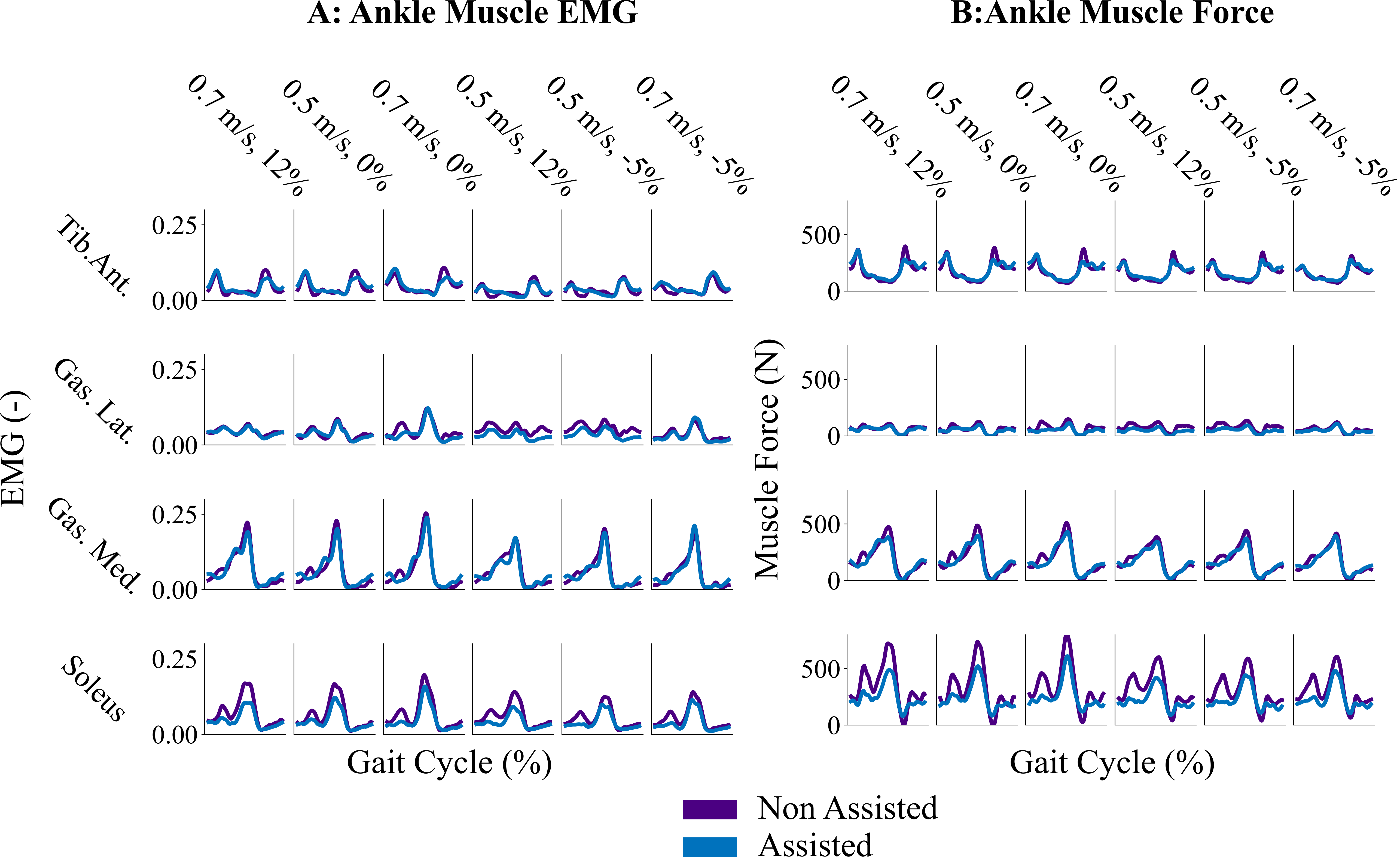}
\caption{Averaged EMG (A) and muscle force (B) for all subjects. Presented for all muscle and all walking speeds and elevations.}
\label{fig_EMG_sup}
\end{figure}

\begin{figure}[h]
\centering
\includegraphics[width=0.4\textwidth]{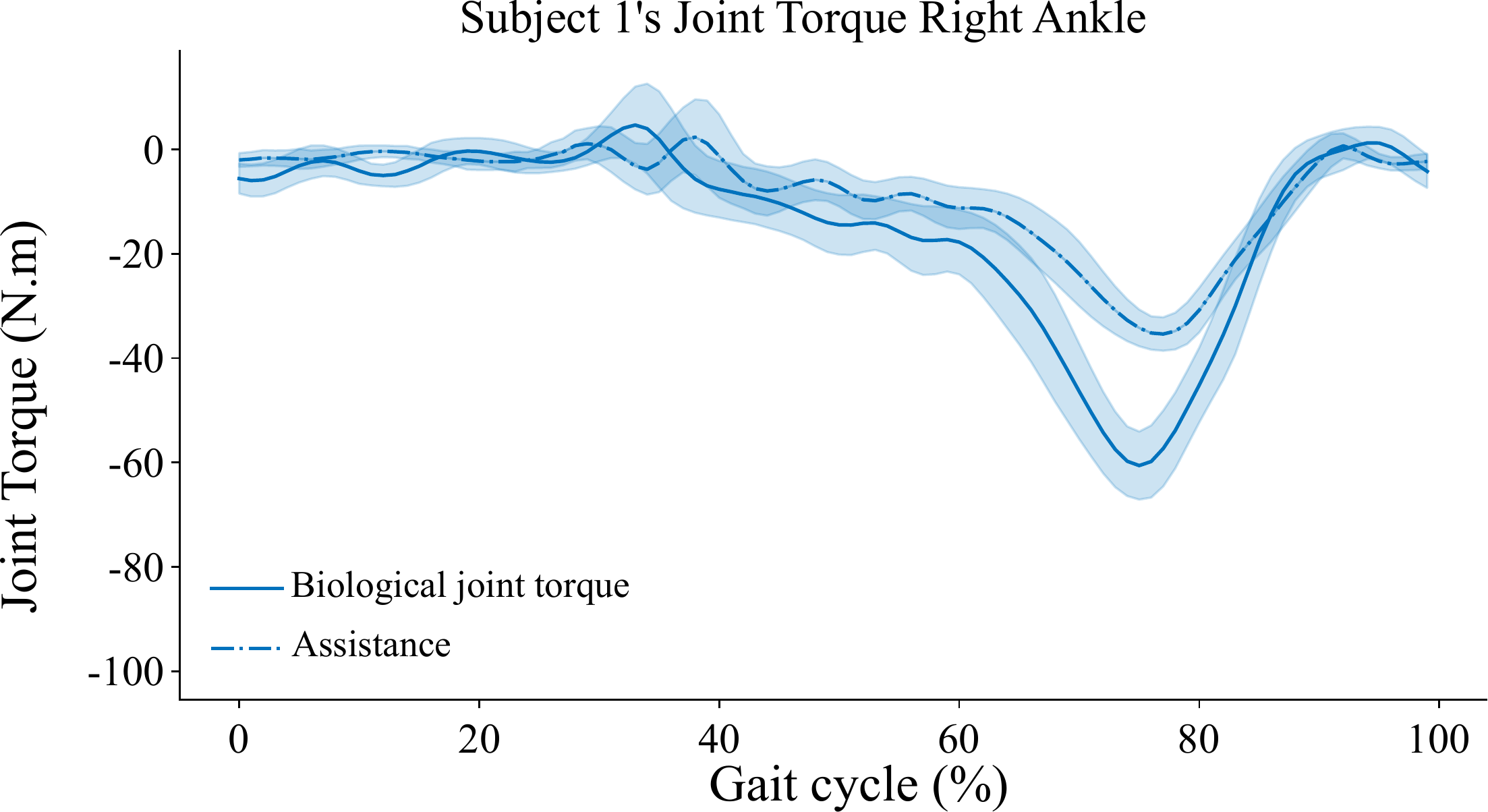}
\caption{Joint torque profiles over the gait cycle for the right side of subject 1. The continuous blue line represents the biological joint torque and the dashed blue line is the assistance provided.}
\label{fig_sbj1}
\end{figure}

\printbibliography

 

\vspace{-33pt}
\begin{IEEEbiography}[{\includegraphics[width=1in,height=1.25in,clip,keepaspectratio]{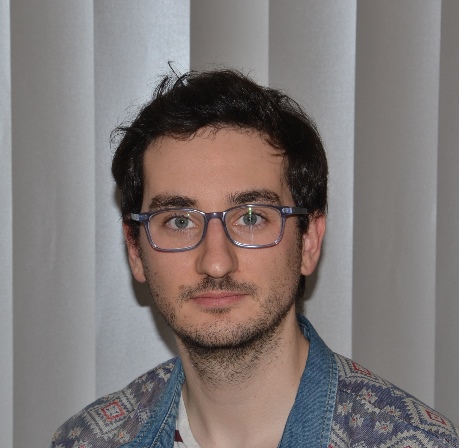}}]{Guillaume Durandau} (Member, IEEE) received the Ph.D. degree from the Department of Biomechanical Engineering at the University of Twente, the Netherlands. He is currently a postdoctoral researcher in the same department and university. Guillaume Durandau's work focuses on interfacing wearable devices with real-time neuromusculoskeletal model.
\end{IEEEbiography}

\vspace{-33pt}
\begin{IEEEbiography}[{\includegraphics[width=1in,height=1.25in,clip,keepaspectratio]{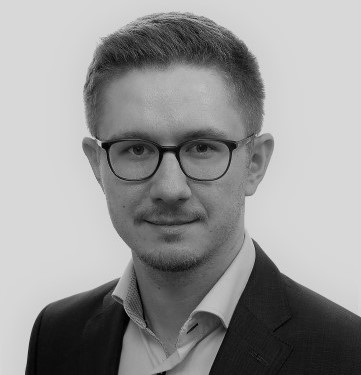}}]{Wolfgang~F.~Rampeltshammer} (Student Member, IEEE) was a PhD student at the Department of Biomechanical Engineering at the University of Twente, the Netherlands. He received his MSc in Electrical Engineering and Information Technology with high distinction in 2016 from the Technical University Munich, Germany. His research focus is the control, high level as well as low level, of assistive robots, especially lower limb exoskeletons, and their optimal interaction with humans by utilizing human-in-the-loop methods.
\end{IEEEbiography}

\vspace{-33pt}
\begin{IEEEbiography}[{\includegraphics[width=1in,height=1.25in,clip,keepaspectratio]{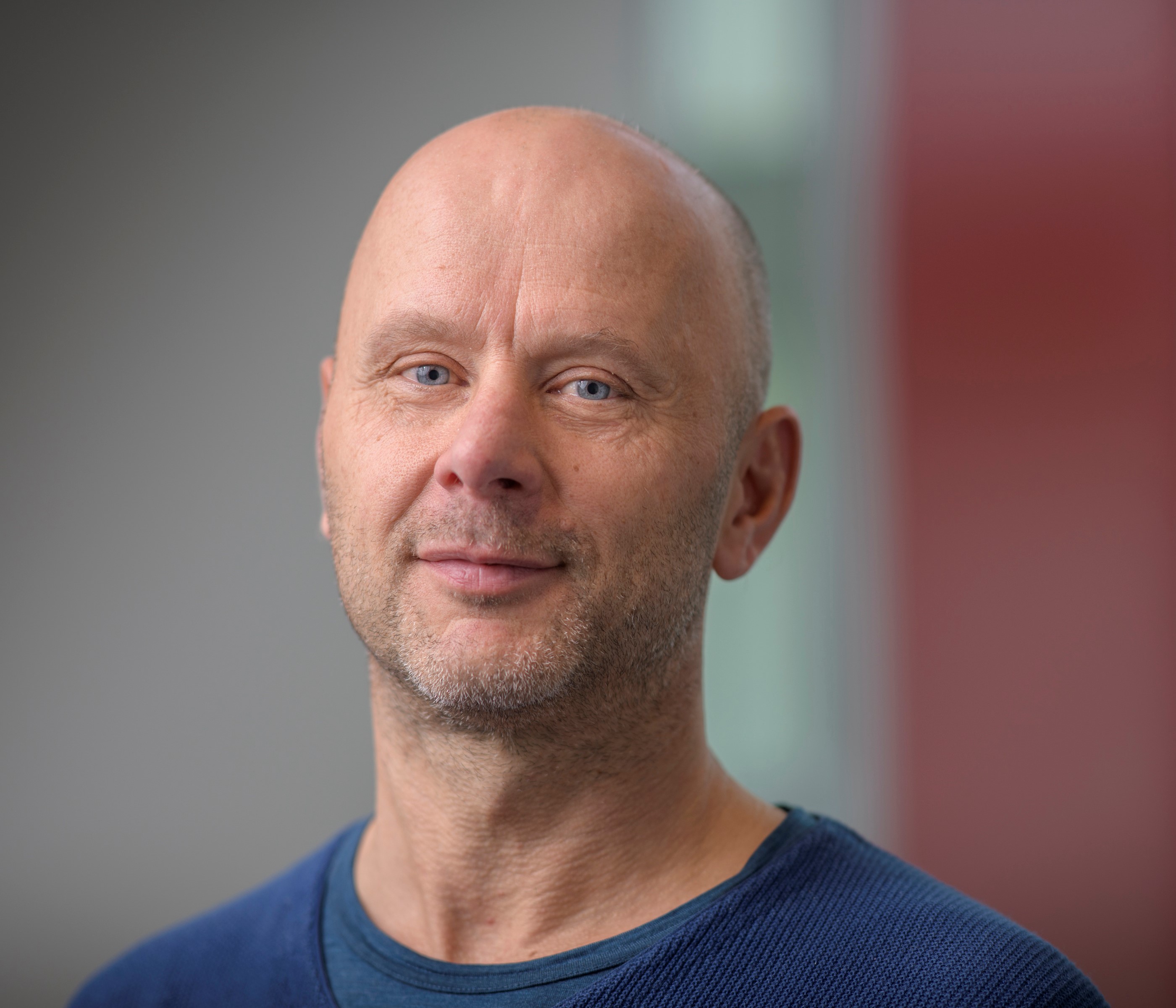}}]{Herman van der Kooij} (Member, IEEE) received the Ph.D. degree with honors (cum laude) from the University of Twente, Enschede, The Netherlands, in 2000. Since 2010, he has been a full Professor in biomechatronics and rehabilitation technology at the Department of Biomechanical Engineering at the University of Twente, and Delft University of Technology, The Netherlands.
\end{IEEEbiography}

\vspace{-33pt}
\begin{IEEEbiography}[{\includegraphics[width=1in,height=1.25in,clip,keepaspectratio]{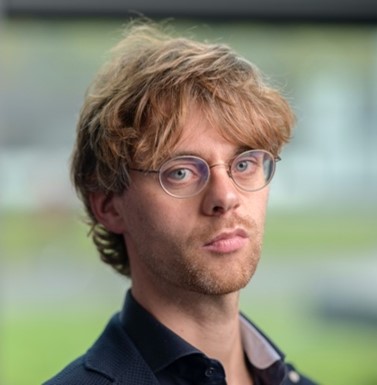}}]{Massimo Sartori}
(Member, IEEE) received the Ph.D. degree in Information Engineering (2011) from the University of Padova, Italy. He was a Postdoc Fellow and Visiting Scholar at University Medical Center Goettingen, Germany, Griffith University, Australia, and Stanford University, USA. He is Professor and Chair in Neuromechanical Engineering at the University of Twente, Netherlands, where he directs the Neuromechanical Modelling and Engineering Lab. 
His work focuses on interfacing robotic technologies with the human neuromuscular system. On these topics he was the recipient of the European Research Council Starting Grant (2018) and Horizon-2020 Projects.
\end{IEEEbiography}


\end{document}